\documentclass[sigconf]{acmart}

\usepackage{booktabs} 

\usepackage{graphicx,caption,subcaption}
\usepackage{amsmath,amssymb,stmaryrd,mathtools, bbm}
\usepackage[ruled,vlined,titlenotnumbered,linesnumbered]{algorithm2e} 
\usepackage[noend]{algpseudocode}
\usepackage{acronym}
\usepackage{comment}
\usepackage{color}
\usepackage{units}
\usepackage{scalerel}
\usepackage{amssymb}

\usepackage[capitalise]{cleveref}

\DeclareCaptionLabelSeparator{periodspace}{.\quad}
\crefname{equation}{}{} 
\crefname{section}{Sec.}{Sec.}

\newcommand{\sys}{\mathcal{S}}
\newcommand{\abs}{\mathcal{M}}

\newcommand{\Env}{\mathcal{E}}
\newcommand{\env}{e}
\newcommand{\envParams}{\mathcal{P}}

\newcommand{\initStates}{\mathcal{X}_0}
\newcommand{\avoidSets}{\mathcal{A}}
\newcommand{\reachSets}{\mathcal{R}}
\newcommand{\avoidSet}{A}
\newcommand{\reachSet}{R}
\newcommand{\signdist}{h}

\newcommand{\dist}{d}
\newcommand{\distMetricOld}{\dist^{a}}
\newcommand{\distMetricNew}{\dist^{b}}
\newcommand{\distMetricEst}{\hat{\dist}_{\epsilon}}

\newcommand{\distMetricIntOne}{\dist^{a2}}

\newcommand{\timeHorizon}{\mathcal{T}}
\newcommand{\horizon}{H}
\renewcommand{\time}{t}

\renewcommand{\state}{x}

\newcommand{\controlspace}{\mathcal{U}}
\newcommand{\controlscheme}{\controlspace_{\Pi}}
\newcommand{\traj}{\xi}

\newcommand{\spec}{\varphi}
\newcommand{\ctrl}{u}

\newcommand{\ctrlseq}{\bold{\ctrl}}
\newcommand{\ctrlseqnull}{\ctrlseq_{\phi}}

\newcommand\given[1][]{\:#1\vert\:}

\newcommand{\reals}{\mathbb{R}}

\newcommand{\domain}{\mathcal{D}}

\newcommand{\prob}{\mathbb{P}}

\newcommand{\metName}{SPEC}
\newcommand{\oldName}{SSM}

\newtheorem{remark}{Remark}

\newtheorem{theorem}{Theorem}
\newtheorem{proposition}{Proposition}

\newtheorem{corollary}{Corollary}
\newtheorem{assumption}{Assumption}

\begin{document}


\title[A New Simulation Metric to Determine Safe Environments and Controllers]{A New Simulation Metric to Determine Safe Environments and Controllers for Systems with Unknown Dynamics}


\author{Shromona Ghosh}
\authornote{Both authors contributed equally to this research. \\ 
** All authors are with the Department of Electrical Engineering and Computer Sciences, University of California, Berkeley. \\
** This research is supported in part by NSF under the CPS Frontier project VeHICaL project (1545126), by NSF grants 1739816 and 1837132, by the UC-Philippine-California Advanced Research Institute under project IIID-2016-005, by the Army Research Laboratory and was accomplished under Cooperative Agreement Number W911NF-17-2-0196, by the DARPA BRASS program under agreement number FA8750-16-C0043, the DARPA Assured Autonomy program under agreement number FA8750-18-C-0101, the iCyPhy center, and by Berkeley Deep Drive.}
\email{shromona.ghosh@berkeley.edu}
\affiliation{
}

\author{Somil Bansal}
\authornotemark[1]
\email{somil@berkeley.edu}
\affiliation{
}

\author{Alberto Sangiovanni-Vincentelli}
\email{alberto@berkeley.edu}
\affiliation{
}

\author{Sanjit A. Seshia}
\email{sseshia@berkeley.edu}
\affiliation{
}

\author{Claire Tomlin}
\email{tomlin@berkeley.edu}
\affiliation{
}

%
\renewcommand{\shortauthors}{Ghosh and Bansal, et al.}

\begin{abstract}
We consider the problem of extracting safe environments and controllers for 
reach-avoid objectives for systems with known state and control spaces, but unknown dynamics.
In a given environment, a common approach is to synthesize a controller from
an abstraction or a model of the system (potentially learned from data). 
However, in many situations, the relationship between the dynamics of the model and the \textit{actual system} is not known; and hence it is difficult to provide safety guarantees for the system.
In such cases, the Standard Simulation Metric (\oldName{}), defined as the worst-case norm distance between the model and the system output trajectories, can be used to modify a reach-avoid specification for the system into a more stringent specification for the abstraction. 
Nevertheless, the obtained distance, and hence the modified specification, can be quite conservative.
This limits the set of environments for which a safe controller can be obtained.
We propose \metName{}, a specification-centric simulation metric, which overcomes these limitations by computing the distance using only the trajectories that violate the specification for the system.
We show that modifying a reach-avoid specification with \metName{} allows us to synthesize a safe controller for a larger set of environments compared to \oldName{}.
We also propose a probabilistic method to compute \metName{} for a general class of systems.
Case studies using simulators for quadrotors and autonomous cars illustrate the advantages of the 
proposed metric for determining safe environment sets and controllers.
\end{abstract}

\setcopyright{none}
\acmISBN{}
\acmDOI{}
\acmYear{}


%

\maketitle

\section{Introduction}
\label{sec:intro}
Recent research in robotics and control theory has focused on developing complex autonomous systems, such as robotic manipulators, autonomous vehicles, and surgical robots. 
Since many of these systems are safety-critical, it is important to design provably-safe controllers while determining environments in
which safety can be guaranteed.
In this work, we focus on reach-avoid objectives, where the goal is to design a controller to reach a target set of states (referred to as reach set) while avoiding unsafe states (avoid set).
Reach-avoid problems are common for autonomous vehicles in the real world; 
for example, a drone flying in an indoor setting. 
Here the reach set could be a desired goal position and the avoid set could be the set of the obstacles.
In such a setting, it is important to determine the environments in which the drone can safely navigate, as well as the corresponding safe controllers. 

Typically, a mathematical model of the system, such as a physics-based first principles model, is used for synthesizing a safe controller in different environments (e.g.,~\cite{tomlin2000game,tabuada2009book}).
However, when the system dynamics are unknown, synthesizing such a controller becomes challenging. 
In such cases, it is a common practice to identify a model for the system.
This model represents an abstraction of the system behavior.
Recently, there has been an increased interest in using machine learning (ML) based tools, such as neural networks and Gaussian processes, for learning abstractions \textit{directly} from the data collected on the system~\cite{bansal2016learning, Bansal2017a, lenz2015deepmpc}. 
One of the many verification challenges for ML-based systems~\cite{seshia-arxiv16} is that such abstractions cannot be directly used for verification, since it is not clear \textit{a priori} how representative the abstraction is of the actual system.
Hence, to use the abstraction to provide guarantees for the system, we need to first quantify the differences between it and the system.

One approach is to use model identification techniques that provide bounds on the mismatch between the dynamics of the system and its abstraction both in time and frequency domains (see \cite{Gevers:2003, hjalmarsson1992estimating, ljung1987system} and references therein).
This bound is then used to design a provably stabilizing controller for the system.
These approaches have largely been limited to linear abstractions and systems, and the focus has been on designing asymptotically stabilizing controllers. 

Another way to quantify the difference between a general non-linear system and its abstraction relies on the notion of a (approximate) \textit{simulation metric} ~\cite{alur2000, girard2007approximate, baier2008principles}. 
Such a metric measures the maximal distance between the system and the abstraction output trajectories over all finite horizon control sequences.
Standard simulation metrics (referred to as \oldName{} here on) 
have been used for a variety of purposes such as safety verification~\cite{Girard2011}, abstraction design for discrete~\cite{larsen1991bisimulation}, nonlinear~\cite{pola2008approximately}, switched~\cite{girard2010approximately} systems, piecewise deterministic and labelled Markov processes~\cite{desharnais2002bisimulation, strubbe2005bisimulation}, and stochastic hybrid systems~\cite{Abate2011, garatti2012simulation, julius2009approximations, bujorianu2005bisimulation}, model checking~\cite{baier2008principles, katoen2007bisimulation}, and model reduction~\cite{dean1997model, papadopoulos2016model}.
\begin{figure}
    \centering
    \includegraphics[scale=0.4]{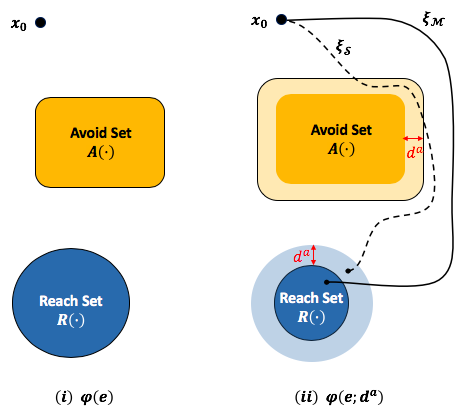}
    \caption{The avoid set is expanded and the reach set is contracted with the simulation metric $\distMetricOld$.
    If the abstraction trajectory ($\traj_{\abs}$) stays clear of the expanded avoid set and reaches the contracted reach set, the system trajectory ($\traj_{\sys}$) also stays clear of the original avoid set and reaches the original reach set.}
    \label{fig:expand_sets}
\end{figure}
Once computed, the \oldName{} is used to \textit{expand} the unsafe set (or avoid set) in \cite{Abate2011}. 
For reach-avoid scenarios, we additionally use it to \textit{contract} the reach set as shown in Figure~\ref{fig:expand_sets}.
If we can synthesize a safe controller that ensures the abstraction trajectory avoids the expanded avoid set and reaches the contracted reach set, then the system trajectory is guaranteed to avoid and reach the original avoid set and reach set respectively. 
This follows from the property that \oldName{} captures the worst case distance between the trajectories of the system and the abstraction. 
Consequently, the set of safe environments for the system can be obtained by finding the set of environments for which we can design a safe controller for the abstraction with the modified specification. 

Even though powerful in its approach, \oldName{} computes the maximal distance between the system and the abstraction trajectories across all possible controllers.
We show in this paper that this is unnecessary and might lead to a conservative bound on the quality of the abstraction for the purposes of controller synthesis.
In particular, the larger the distance between the system and the abstraction, the larger the expansion (contraction) of the avoid (reach) set. 
In many cases, this results in unrealizability wherein there does not exist a safe controller for the abstraction for the modified specification. 

In this paper, we propose \metName{}, SPEcification-Centric simulation metric, that overcomes these limitations.
\metName{} achieves this by computing the distance across %
\begin{enumerate}
    \item only those controllers that can be synthesized by a particular control scheme and that are safe for the abstraction (in the context of the original reach-avoid specification) --- these are the only potential safe controllers for the system;
    \item only those abstraction and system trajectories for which the system violates the reach-avoid specification, and
    \item only between the abstraction trajectory and the reach and the avoid sets.
\end{enumerate}
%
If the reach-avoid specification is changed using \metName{} in a similar fashion as that for \oldName{}, it is guaranteed that if a controller is safe for the abstract model, it remains 
safe for the system.
\metName{} can be significantly less conservative than \oldName{}, and can be used to design safe controllers for the system for a broader range of reach-avoid specifications. 
In fact, we show that, among all uniform distance bounds (i.e., a single distance bound is used to modify the specification in all environments), \metName{} provides the largest set of environments such that a safe controller for the abstraction is also safe for the system.

Note that a similar metric has been used earlier~\cite{Ghosh:2016} to find tight environment assumptions for temporal logic specifications. 
However, it applies in much more restricted settings since it 
relies on having simple linear representations of the abstraction 
which can be expressed as a linear optimization problem.

In general, it is challenging to compute both \oldName{} and \metName{} when the dynamics of the system are not available. 
Several approaches have been proposed in the literature for computing \oldName{}~\cite{abate2009contractivity, girard2007approximate, julius2009approximations}; however, restrictive assumptions on the dynamics of the systems are often required to compute it. 
More recently, a randomized approach has been proposed to compute \oldName{}~\cite{Abate2011, garatti2012simulation} for finite-horizon properties that relies on ``scenario optimization'', which was first introduced for solving robust convex programs via randomization~\cite{calafiore2005uncertain} and then extended to semi-infinite chance-constrained optimization problems~\cite{campi2011}. 
Scenario optimization is a sampling-based method to solve semi-infinite optimization problems, and has been used for system and control design~\cite{calafiore2006, campi2009}. 
In this work, we propose a scenario optimization-based computational method for \metName{} that has general applicability and is not restricted to a specific class of systems. 
Indeed, the only assumption is that the system is available as an oracle, with known state and control spaces, which we can simulate to determine the corresponding output trajectory.
Given that the distance metric is obtained via randomization and, hence, is a random quantity, we provide probabilistic guarantees on the performance of \metName{}.
However, this confidence is a design parameter and can be chosen as close to 1 as desired (within a simulation budget).
To summarize, this paper's main contributions are:
\begin{itemize}
    \item \metName{}, a new simulation metric that is less conservative than \oldName{}, and provides the largest set of environments such that a safe controller for the abstraction is also safe for the system;
    \item a method to compute \metName{} that is not restricted to a specific class of systems, and 
    \item a demonstration of the proposed approach on numerical examples and simulations of real-world autonomous systems, such as a quadrotor and an autonomous car.
\end{itemize}


\section{Mathematical Preliminaries}
Let $\sys$ be an unknown, discrete-time, potentially non-linear, dynamical system with state space $\mathbb{R}^{n_x}$ and control space $\mathbb{R}^{n_u}$. 
Let $\abs$ be an abstraction of $\sys$ with the same state and control spaces as $\sys$, whose dynamics are known. 
We also assume that the bounds between the dynamics of $\abs$ and $\sys$ are not available beforehand (i.e., we cannot \textit{a priori} quantify how different the two are).
$\traj_{\sys}(\time; \state_0, \ctrlseq)$ denotes the trajectory of $\sys$ at time $t$ starting from the initial state $\state_0$ and applying the controller $\ctrlseq$.
$\traj_{\abs}$ is similarly defined. 
For ease of notation, we drop $\ctrlseq$ and $\state_0$ from the trajectory arguments wherever convenient.

We define by $\Env := \initStates \times \avoidSets \times \reachSets$ the set of all reach-avoid scenarios (also referred to as \textit{environment scenarios} here on), for which we want to synthesize a controller for $\sys$.
A reach-avoid scenario $\env \in \Env$ is a three-tuple, $(\state_0, \avoidSet(\cdot), \reachSet(\cdot))$, where $\state_0 \in \initStates \subset \mathbb{R}^{n_x}$ is the initial state of $\sys$.
$\avoidSet(\cdot) \in \avoidSets, \avoidSet(\cdot) \subset \mathbb{R}^{n_x}$ and $\reachSet(\cdot) \in \reachSets, \reachSet(\cdot) \subset \mathbb{R}^{n_x}$ are (potentially time varying) sequences of avoid and reach sets respectively.
We leave $\avoidSets$ and $\reachSets$ abstract except where necessary.
If the sets are not time varying, we can replace $\reachSet(\cdot)$ (respectively $\avoidSet(\cdot)$) by the stationary $\reachSet$ (respectively $\avoidSet$).
Similarly, if there is no avoid or reach set at a particular time, we can represent $\avoidSet(\time) = \emptyset$ and $\reachSet(\time) = \mathbb{R}^{n_x}$.

For each $\env \in \Env$, we define a reach-avoid specification, $\spec(\env)$ 
\begin{equation}
    \label{eqn:spec:unmodified}
    \spec(\env) :=\{\traj(\cdot): 
    \; \forall t \in \timeHorizon   \;
    \traj(t) \notin \avoidSet(t) \wedge \traj(t) \in \reachSet(t) 
    \},
\end{equation}
where $\timeHorizon$ denote the time-horizon $\{0, 1, \ldots, \horizon\}$.
We say $\traj(\cdot)$ satisfies the specification $\spec(\env)$, denoted $\traj(\cdot)\models \spec(\env)$, if $\traj(\cdot)\in \spec(\env)$.

The reader might observe that our use of $\reachSet(\cdot)$ in~\eqref{eqn:spec:unmodified} differs somewhat from the intuitive notion
of a reach set (depicted, e.g., in Fig.~\ref{fig:expand_sets}). 
Specifically, \eqref{eqn:spec:unmodified} defines the reach-avoid specification such that the output trajectory must remain within $\reachSet(t)$ at all times $t$, while the usual notion involves \textit{eventually} reaching a desired set of states. Note, however, that for the purposes of defining $\spec(\env)$, these notions are equivalent if $\reachSet(t)$ in~\eqref{eqn:spec:unmodified} represents the backwards reachable tube corresponding to the desired reach set: if a state is reachable eventually,
then the trajectory stays within the backwards reachable tube at all time points. We henceforth use the $\reachSet(t)$ in the latter sense since it simplifies the mathematics in the paper.

Finally, we define $\controlscheme(\env) \subset \controlspace$ to be the space of all permissible controllers for $\env$, and $\controlspace$ to be the space of all finite horizon control sequences over $\timeHorizon$.
For example, if we restrict ourselves to linear feedback controllers, $\controlscheme$ represents the set of all linear feedback controllers that are defined over the time horizon $\timeHorizon$.
\section{Problem Formulation}
Given the set of reach-avoid scenarios $\Env$, the controller scheme $\controlscheme$, and the abstraction $\abs$, our goal is two-fold:
\begin{enumerate}
    \item to find the environment scenarios for which it is possible to design a controller such that $\traj_{\sys}(\cdot)$ satisfies the corresponding reach-avoid specification $\spec(\env)$,
    \item to find a corresponding safe controller for each scenario in (1).
\end{enumerate}
Mathematically, we are interested in computing the set $\Env_{\sys}$  
\begin{equation}
\label{eqn:ideal_sets:1}
\Env_{\sys} =  \{\env \in \Env \,: \exists \ctrlseq \in \controlscheme(\env)\;, \traj_{\sys}(\cdot; \state_0, \ctrlseq) \models \spec(\env) \},
\end{equation}
and the corresponding set of safe controllers $\controlspace_{\sys}(\env)$ for each $\env \in \Env_{\sys}$
\begin{equation}
\label{eqn:ideal_sets:2}
\controlspace_{\sys}(\env) =  \{\ctrlseq \in \controlscheme(\env)\;: \traj_{\sys}(\cdot; \state_0, \ctrlseq) \models \spec(\env) \}.
\end{equation}
When a dynamics model of $\sys$ is known, several methods have been studied in literature to compute the sets $\Env_{\sys}$ and $\controlspace_{\sys}(\env)$ for reach-avoid problems~\cite{tomlin1998conflict, mitsch2013provably, tomlin2000game}.
However, since a dynamics model of $\sys$ is unknown, the computation of these sets is challenging in general. 
To overcome this problem, one generally relies on the abstraction $\abs$. 
We make the following assumptions on $\sys$ and $\abs$:
\begin{assumption}
$\sys$ is available as an oracle that can be simulated, i.e.,
we can run an execution (or experiment) on $\sys$ and obtain the corresponding system trajectory $\traj_{\sys}(\cdot)$. 
\end{assumption}
\begin{assumption}
For any $\env \in \Env$, we can determine if there exist a controller such that $\traj_{\abs} \models \spec(\env)$ and can compute such a controller.
\end{assumption}
Assumption 1 states that even though we do not know the dynamics of $\sys$, we can run an execution of $\sys$. 
Assumption 2 states that it is possible to verify whether $\abs$ satisfies a given specification $\spec(\env)$ or not.
Although it is not a straightforward problem, since the dynamics of $\abs$ are known, several existing methods can be used for obtaining a safe controller for $\abs$.  

Under these assumptions, we show that we can convert a verification problem on $\sys$ to a verification problem on $\abs$. 
In particular, we compute a distance bound, \metName, between $\sys$ and $\abs$ which along with $\abs$ allows us to compute a conservative approximation of $\Env_{\sys}$ and $\controlspace_{\sys}(\env)$.

\section{Running Example} \label{sec:running_example:description}
We now introduce a very simple example that we will use to illustrate our approach, a 2 state linear system in which the system and the abstraction differ only in one parameter. 
Although simple, this example illustrates several facets of \metName{}.
We present more realistic case studies in Section~\ref{sec:case_studies}.

Consider a system $\sys$ whose dynamics are given as
\begin{equation} \label{eqn:running_example:sys_dynamics}
\state(\time+1) = \begin{bmatrix} \state_1(\time+1) \\ \state_2(\time+1) \end{bmatrix} = \begin{bmatrix} 2 & 0 \\ 0 & 0.1 \end{bmatrix}\begin{bmatrix} \state_1(\time) \\ \state_2(\time) \end{bmatrix} +  \begin{bmatrix} 1 \\ 0 \end{bmatrix}\ctrl(\time).
\end{equation}
We are interested in designing a controller for $\sys$ to regulate it from the initial state $\state(0) := \state_0 = \left[0, 0\right]$ to a desired state $\state^* = \left[\state_1^*, 0\right]$ over a time-horizon of 20 steps, i.e, $\horizon=20$. In particular, we have
\begin{equation*}
\initStates = \{ \left[0, 0\right] \},\quad \avoidSets = \emptyset,\quad \reachSets = \bigcup_{-4 \leq \state_1^* \leq 4} \reachSet(\cdot; \state^*),
\end{equation*}
where 
\begin{align*} 
\reachSet(\time; \state^*) = & \mathbb{R}^2, \time \in \{0, 1, \ldots, \horizon-1 \}, \nonumber \\
\reachSet(\horizon; \state^*) = &\{\state: \|\state - \state^*\|_2 < \gamma\}.
\end{align*}
We use $\gamma = 0.5$ in our simulations.
Thus, each $\env \in \Env$ consists of a final state $\state^*$ (equivalently, a reach set $\reachSet(\horizon; \state^*)$) to which we want the system to regulate, starting from the origin.
Consequently, the system trajectory satisfies the reach-avoid specification in this case if $\traj_{\sys}(\horizon; \state_0, \ctrlseq) \in \reachSet(\horizon; \state^*)$.

For the purpose of this example, we assume that the system dynamics in \eqref{eqn:running_example:sys_dynamics} are unknown; only the dynamics of its abstraction $\abs$ are known and given as
\begin{equation} \label{eqn:running_example:abs_dynamics}
\state(\time+1) = \begin{bmatrix} \state_1(\time+1) \\ \state_2(\time+1) \end{bmatrix} = \begin{bmatrix} 2 & 0 \\ 0 & 0.1 \end{bmatrix}\begin{bmatrix} \state_1(\time) \\ \state_2(\time) \end{bmatrix} +  \begin{bmatrix} 1 \\ 0.1 \end{bmatrix}\ctrl(\time).
\end{equation}

In this example, we use the class of linear feedback controllers as $\controlscheme(\env)$, although other control schemes can very well be used. In particular, for any given environmental scenario $\env$, the space of controllers $\controlscheme(\env)$ is given by
\begin{equation*}
\controlscheme(\env) = \{LQR(q, \state^*): 0.1 \leq q \leq 100\},
\end{equation*}
where $LQR(q, \state^*)$ is a Linear Quadratic Regulator (LQR) designed for the abstraction dynamics in \eqref{eqn:running_example:abs_dynamics} to regulate the abstraction trajectory to $\state^*$ \footnote{That is, we penalize the trajectory deviation to the desired state $\state^*$ in the LQR cost function.}, with the state penalty matrix $Q = qI$ and the control penalty coefficient $R = 1$.
Here, $I \in \mathbb{R}^{2 \times 2}$ is an identity matrix.
Thus, for different values of $q$ we get different controllers, which affect the various characteristics of the resultant trajectory, such as overshoot, undershoot, and final state. 
$LQR(q)$ for any given $q$ can be obtained by solving the discrete-time Riccati equation~\cite{kwakernaak1972linear}.
Our goal thus is to use the dynamics in \eqref{eqn:running_example:abs_dynamics} to find the set of final states to which $\sys$ can be regulated and the corresponding regulator in $\controlscheme(\env)$.

\section{Solution Approach}
\label{sec:approach}
\subsection{Computing approximate safe sets using $\abs$ and simulation metric}
Computing sets $\Env_{\sys}$ and $\controlspace_{\sys}$ exactly can be challenging since the dynamics of $\sys$ are unknown \textit{a priori}. 
Generally, we use the abstraction $\abs$ as a replacement for $\sys$ to synthesize and analyze safe controllers for $\sys$. 
However, to provide guarantees on $\sys$ using $\abs$, we would need to quantify how different the two are.
 
We quantify this difference through a distance bound, $\dist$, between $\sys$ and $\abs$.
$\dist$ is used to modify the specification $\spec(\env)$ to a more stringent specification $\spec(\env; \dist)$ such that if $\traj_{\abs}(\cdot) \models \spec(\env; \dist)$ then $\traj_{\sys}(\cdot) \models \spec(\env)$.
Thus, the set of safe controllers for $\abs$ for $\spec(\env; \dist)$ can be used as an approximation for $\controlspace_{\sys}(\env)$.
In particular, if we define the sets $\controlspace_{\spec(\env; \dist)}$ and $\Env_{\spec}(\dist)$ as 
\begin{equation}
\begin{split}
\label{eqn:controller:modified}
        \controlspace_{\spec(\env; \dist)}  & :=\{\ctrlseq \in \controlscheme(\env): \traj_{\abs}(\cdot; \state_0, \ctrlseq) \models \spec(\env; \dist)\} \\
        \Env_{\spec}(\dist) & := \{\env \in \Env: \controlspace_{\spec(\env, d)} \neq \emptyset\},
\end{split}
\end{equation}
then $\controlspace_{\spec(\env; \dist)}$ and $\Env_{\spec}(\dist)$ can be used as an approximation of $\controlspace_{\sys}(\env)$ and $\Env_{\sys}$ respectively.
Consequently, a verification problem on $\sys$ can be converted into a verification problem on $\abs$ using the modified specification.

One such distance bound $\dist$ is given by the simulation metric, \oldName{}, between $\abs$ and $\sys$ defined as
\begin{equation}
\label{eqn:sim_metric:naive}
    \distMetricOld = \max_{\env \in \Env} \max_{\ctrlseq \in \controlscheme(\env)} \|\traj_{\sys}(\cdot; \state_0, \ctrlseq) - \traj_{\abs}(\cdot; \state_0, \ctrlseq) \|_{\infty}
\end{equation}
Here, the $\infty$-norm is the maximum distance between the trajectories across all timesteps.
Typically \oldName{} is computed over the space of all finite horizon controls $\controlspace$ instead of $\controlscheme(\env)$~\cite{Girard2011}. 
Since we are interested in a given control scheme, we restrict this computation to $\controlscheme(\env)$.
In general, $\distMetricOld$ is difficult to compute, because it requires searching over (the potentially infinite) space of controllers and environments. 
An approximate technique to compute $\distMetricOld$ was presented for systems whose dynamics were unknown with probabilistic guarantees in~\cite{Abate2011}.

However, if $\distMetricOld$ can be computed then it can be used to modify a specification $\spec(\env)$ to $\spec(\env; \distMetricOld)$ as follows: ``expand'' the avoid set $\avoidSet(\cdot)$ to get the augmented avoid set $\avoidSet(\cdot; \distMetricOld) = \avoidSet(\cdot) \oplus \distMetricOld$, and ``contract'' the reach set $\reachSet(\cdot)$ to obtain  a conservative reach set $\reachSet(\cdot; \distMetricOld) = \reachSet(\cdot) \ominus \distMetricOld$ (see Figure~\ref{fig:expand_sets}). 
Here, $\oplus$ ($\ominus$) is the Minkowski sum(difference)\footnote{The Minkowski sum of a set $K$ and a scalar $d$ is the set of all points that are the sum of any point in $K$ and $B(d)$, where $B(d)$ is a disc of radius $d$ around the origin.}. Consequently, $\spec(\env; \distMetricOld)$ is the set of trajectories which avoid $\avoidSet(\cdot; \distMetricOld)$ and are always contained in $\reachSet(\cdot; \distMetricOld)$, 
\begin{equation}
    \label{eqn:spec:modified}
    \spec(\env; \distMetricOld) :=\{\traj(\cdot): \traj(t) \notin \avoidSet(t; \distMetricOld), \traj(t) \in \reachSet(t; \distMetricOld) \forall t \in \timeHorizon\}.
\end{equation}
Then it can be shown that any controller that satisfies the specification $\spec(\env; \distMetricOld)$ for $\abs$ also ensures that $\sys$ satisfies the specification $\spec(\env)$.
\begin{proposition} \label{prop1}
For any $\env \in \Env$ and controller $\ctrlseq \in \controlscheme(\env)$, we have $\traj_{\abs}(\cdot; \state_0, \ctrlseq) \models \spec(\env; \distMetricOld)$ implies $\traj_{\sys}(\cdot; \state_0, \ctrlseq) \models \spec(\env)$.
\end{proposition}
\noindent The proof of Proposition \ref{prop1} can be found in the Appendix.
Proposition~\ref{prop1} implies that $\Env_{\spec}(\distMetricOld)$ and $\controlspace_{\spec(\env; \distMetricOld)}$ can be used as approximations of $\Env_{\sys}$ and $\controlspace_{\sys}(\env)$ respectively.
However, the distance bound in~\eqref{eqn:sim_metric:naive} does not take into account the reach-avoid specification (environment) for which a controller needs to be synthesized. 
Thus, $\distMetricOld$ can be quite conservative. 
As a result, the modified specification can be so stringent that the set of environments $\Env_{\spec}(\distMetricOld)$ for which we can synthesize a provably safe controller for the abstraction (and hence for the system) itself will be very small, resulting in a very conservative approximation of $\Env_{\sys}$.

\subsection{Specification-Centric Simulation Metric (\metName)}
To overcome these limitations, we propose \metName,
%
\begin{equation}
\label{eqn:sim_metric:proposed}
    \distMetricNew = \max_{\env \in \Env} \max_{u \in \controlspace_{\spec(\env)}}
    \dist(\traj_{\sys}(\cdot),\traj_{\abs}(\cdot)),
\end{equation}

%
where
\begin{equation}
    \label{eqn:dist_fn:proposed}
\begin{split}
    \dist(\traj_{\sys}(\cdot), \traj_{\abs}(\cdot)) = \min_{\time \in \timeHorizon}
    (\min\{& \signdist\left(\traj_{\abs}(\time; \state_0, \ctrlseq), \avoidSet(\time)\right), \\ &
    -\signdist\left(\traj_{\abs}(\time; \state_0, \ctrlseq), \reachSet(\time)\right)
    \}) \mathbbm{1}_{\left(\traj_{\sys}(\cdot) \not\models \spec(\env)\right)}
\end{split}
\end{equation}
%
%
%

Here $\controlspace_{\spec(\env)}:= \{\ctrlseq \in \controlspace_{\Pi}(\env): \traj_{\abs}(\cdot; \state_0, \ctrlseq) \models \spec(\env)\}$ is the set of all controls such that $\abs$ satisfies the specification $\spec(\env)$. 
%
%
$\mathbbm{1}_l$ represents the indicator function which is $1$ if $l$ is true and $0$ otherwise, 
$\signdist(\state, K)$ is the signed distance function defined as
\begin{equation*}
    \signdist(\state, K) := 
\begin{cases}
    \inf_{k \in K} \|\state - k\|, & \text{if } x\not\in K\\
    -\inf_{k \in K^C} \|\state - k\|, & \text{otherwise.}
\end{cases}
\end{equation*}
\noindent If for any $\env \in \Env$, $\controlspace_{\spec(\env)}$ is empty, we define the distance function $\dist(\traj_{\sys}(\cdot), \traj_{\abs}(\cdot))$ to be zero. 
Similarly, if there is no $\avoidSet(\cdot)$ or $\reachSet(\cdot)$ at a particular $\time$, the corresponding signed distance function is defined to be $\infty$.
%
There are four major differences between \eqref{eqn:sim_metric:naive} and \eqref{eqn:sim_metric:proposed}:
\begin{enumerate}
    \item To compute the $\distMetricNew$ we only consider the feasible set of controllers that can be synthesized by the control policy, $\controlspace_{\spec(\env)} \subseteq \controlscheme(\env)$, as all other controllers do not help us in synthesizing a safe controller for $\sys$ (as they are not even safe for $\abs$).
    \item To compute the distance between $\sys$ and $\abs$, we only consider those trajectories where $\sys$ violates the specification.
    This is because a non-zero distance between the trajectories of $\sys$ and $\abs$, where the $\traj_{\sys} \models \spec(\env)$ does not give us any additional information in synthesizing a safe controller.
    \item Within a falsifying $\traj_{\sys}$, we compute the minimum distance of the abstraction trajectory from the avoid and reach sets rather than the system trajectory, as that is sufficient to obtain a margin to discard behaviors that are safe for the abstraction but unsafe for the system.
    \item Finally, a minimum over time of this distance is sufficient to discard an unsafe trajectory, as the trajectory will be unsafe if it is unsafe at any $\time$.
\end{enumerate}
These considerations ensure that $\distMetricNew$ is far less conservative compared to $\distMetricOld$ and allows us to synthesize a safe controller for the system for a wider set of environments.
We first prove that $\distMetricNew$ can be used to compute an approximation of $\Env_{\sys}$.
\begin{proposition} \label{prop:main_result_1}
If $\; \controlspace_{\spec(\env; \distMetricNew)} \subseteq \controlspace_{\spec(\env)}$, then $\traj_{\abs}(\cdot; \state_0, \ctrlseq) \models \spec(\env; \distMetricNew)$ implies $\traj_{\sys}(\cdot; \state_0, \ctrlseq) \models \spec(\env)\; \forall \env \in \Env, \ctrlseq \in \controlscheme(\env).$
\end{proposition} 
\noindent The proof of Proposition \ref{prop:main_result_1} can be found in the Appendix. 
Thus, if we define $\controlspace_{\spec(\env; \distMetricNew)}$ and $\Env_{\spec}(\distMetricNew)$ as in \eqref{eqn:controller:modified} then they can be used as approximations of $\controlspace_{\sys}(\env)$ and $\Env_{\sys}$ respectively.
Proposition~\ref{prop:main_result_1} requires that the set of controllers that satisfy the modified specification, $\controlspace_{\spec(\env; \distMetricNew)}$, is a subset of the set of the controllers that satisfy the actual specification, $\controlspace_{\spec(\env)}$. 
When $\controlscheme(\env)=\controlspace$, this condition is trivially satisfied as the modified specification is more stringent than the actual specification.
Other control schemes, such as the set of linear feedback controllers and feasibility-based optimization schemes also satisfy this condition.
In fact, in such cases, the proposed metric, $\distMetricNew$, quantifies the tightest (largest) approximation of $\Env_{\sys}$, i.e., $\nexists d < \distMetricNew$, such that $\Env_{\spec}(d) \subseteq \Env_{\sys}$.
%
\begin{theorem}
\label{thm:min_dist}
Let $\controlscheme$ be such that $\controlspace_{\spec(\env; d_1)} \subseteq \controlspace_{\spec(\env; d_2)}$ whenever $d_1 > d_2$. 
Let $\dist \in \mathbb{R}^{+}$ be any distance bound such that 
\begin{equation} \label{eqn:theorem1:condition}
    \forall \env \in \Env\;, \forall \ctrlseq \in \controlscheme(\env)\;, \traj_{\abs}(\cdot) \models \spec(\env; \dist) \rightarrow \traj_{\sys}(\cdot) \models \spec(\env).
\end{equation}
Then $\forall \env \in \Env, \controlspace_{\spec(\env; \dist)} \subseteq \controlspace_{\spec(\env; \distMetricNew)} \subseteq \controlspace_{\sys}(\env)$. 
Moreover, $\Env_{\spec}(\dist) \subseteq \Env_{\spec}(\distMetricNew) \subseteq \Env_{\sys}$. 
Hence, $\Env_{\spec}(\distMetricNew)$ and $\controlspace_{\spec(\env; \distMetricNew)}$ quantify the tightest (largest) approximations of $\Env_{\sys}$ and $\controlspace_{\sys}(\env)$ respectively among all uniform distance bounds $d$.
\end{theorem}

\noindent Theorem~\ref{thm:min_dist} states that $\distMetricNew$ is the smallest among all (uniform) distance bounds between $\abs$ and $\sys$, such that a safe controller synthesized on $\abs$ is also safe for $\sys$. 
Even though this is a stricter condition than we need for defining $\Env_{\sys}$, where we care about the existence of at least one safe controller for $\sys$, it allows us to use \textit{any}  safe controller for $\abs$ as a safe controller for $\sys$.
Formally, $\distMetricNew \leq \dist$, for all $\dist \in \reals^{+}$ such that $\forall \env \in \Env_{\spec}(\dist) \;,\forall \ctrlseq \in \controlspace_{\spec(\env; \dist)}\;, \traj_{\sys}(\cdot) \models \spec(\env)$. 

Intuitively, to compute~\eqref{eqn:sim_metric:proposed}, we collect all $\traj_{\abs}(\cdot), \traj_{\sys}(\cdot)$ pairs (across all $\env \in \Env$ and $\ctrlseq \in \controlspace_{\spec(\env)}$) where $\traj_{\abs}(\cdot) \models \spec(\env)$ and $\traj_{\sys}(\cdot) \not\models \spec(\env)$. 
We then evaluate~\eqref{eqn:dist_fn:proposed} for each pair and take the maximum to compute $\distMetricNew$. By expanding (contracting) every $\avoidSet(\cdot) \in \avoidSets$ ($\reachSet(\cdot) \in\reachSets$) uniformly by $\distMetricNew$, we ensure that none of the $\traj_{\abs}(\cdot)$ collected above is feasible once the specification is modified, and hence, $\traj_{\sys}(\cdot)$ will never falsify $\spec(\env)$.
To ensure this, we prove that $\distMetricNew$ is the minimum distance by which the avoid sets should be augmented (or the reach sets should be contracted). 
Thus, $\distMetricNew$ can also be interpreted as the minimum $\dist$ by which the specification should be modified to ensure that $\controlspace_{\spec(\env; \dist)} \subseteq \controlspace_{\sys}(\env)$ for all $\env \in \Env$.
\begin{corollary}
\label{cor:spec_containment}
Let $\dist \in [0, \distMetricNew]$ satisfies \eqref{eqn:theorem1:condition},
%
%
then $\traj_{\abs}(\cdot) \models \spec(\env; \dist)$ implies $\traj_{\abs}(\cdot) \models \spec(\env; \distMetricNew)$.
\end{corollary}
\noindent We conclude this section by discussing the relative advantages and limitations of \metName{} and \oldName{}, and a few remarks.

\paragraph{\textbf{Comparing \metName{} and \oldName{}}} 
\oldName{} is specification-independent (and hence environment-independent); and hence can be reused across different tasks and environments. This is ensured by computing the distance between trajectories across all input control sequences; however, the very same aspect can make \oldName{} overly-conservative.
Making \metName{} specification-dependent trades in generalizability for a less conservative measure. Although environment-dependent, the set of safe environments obtained using \metName{} is larger compared to \oldName{}. This is an important trade-off to make for any distance metric--the utility of a distance metric could be somewhat limited if it is too conservative. 

The computational complexities for computing \metName{} and \oldName{} are the same since they both can be computed using Algorithm~\ref{alg:solution}. To compute \oldName{} we sample from a domain of all finite horizon controls. To compute \metName{} we additionally need to be able to define and sample from the set of environment scenarios, but we believe that some representation of the environment scenarios is important for practical applications. 
%

\begin{remark}
The proposed framework can also be used in the scenarios where there is a deterministic controller for each environment. In such cases, $\controlscheme(\env)$ (and $\controlspace_{\spec(\env)}$) is a singleton set for every environment $\env$ (see Section \ref{sec:autonomous_car} for an example). 
However, from a control theory perspective, it might be useful to have a set of safe controllers that have different transient behaviors, that the system designer can choose from without recomputing the distance metric.
\end{remark}


\section{Distance Metric Computation}
\label{sec:computation}
Since a dynamics model of $\sys$ is not available, 
the computation of the distance bound $\distMetricNew$ is generally difficult.
Interestingly, this computational issue can be resolved using a randomized approach, such as scenario optimization~\cite{calafiore2006}. 
Scenario optimization has been used for a variety of purposes~\cite{campi2009, campi2011}, such as robust control, model reduction, as well as for the computation of \oldName{}~\cite{Abate2011}.

Computing $\distMetricNew$ by scenario optimization is summarized in Algorithm~\ref{alg:solution}.
We start by (randomly) extracting $N$ realizations of the environment $\env_i$, $i = 1, 2, \ldots, N$ (Line 2).
Each realization $\env_i$ consists of an initial state $\state_0^i$, and a sequence of reach and avoid sets, $\avoidSet^i(\time)$ and $\reachSet^i(\time)$.
For each $\env_i$, we extract a controller $\ctrlseq_i \in \controlspace_{\spec(\env_i)}$ (Line 5). If such a controller does not exist, we denote $\ctrlseq_i$ to be a null controller $\ctrlseqnull$. 
$\ctrlseq_i$ (if not $=\ctrlseqnull$) is then applied to both the system as well as the abstraction to obtain the corresponding trajectories $\traj_{\sys}^i(\cdot; \state_0^i, \ctrlseq_i)$ and $\traj_{\abs}^i(\cdot; \state_0^i, \ctrlseq_i)$ (Line 6).
We next compute the distance between these two trajectories, $\dist_i$, using \eqref{eqn:dist_fn:proposed} (Line 7). If $\ctrlseq_i = \ctrlseqnull$, no satisfying controller exists for $\abs$, and hence $\dist(\traj_{\sys}(\cdot; \state_0^i, \ctrlseq_n), \traj_{\abs}(\cdot; \state_0^i, \ctrlseq_n))$ is trivially $0$.
The maximum across all these distances, $\distMetricEst$, is then used as an estimate for $\distMetricNew$ (Line 10).

Although simple in its approach, scenario optimization provides provable approximation guarantees. In Algorithm~\ref{alg:solution}, we have to sample both an $\env\in \Env$ and a corresponding controller $\ctrlseq \in \controlspace_{\spec(\env)}$. 
We define a joint sample space 
\begin{equation}
\label{eqn:domain}
\domain = \{(\env \times \controlspace_{\spec(\env)}): \env \in \Env, \controlspace_{\spec(\env)} \neq \emptyset \} \cup \{(\env, \ctrlseqnull): \env \in \Env,  \controlspace_{\spec(\env)} = \emptyset\}
\end{equation} 
$\domain$ contains all feasible $(\env, \ctrlseq)$ pairs for $\abs$. We create a dummy sample $(\env, \ctrlseqnull)$ for all $\env$ where a satisfying controller does not exist for the abstraction. 
We next define a probability distribution on $\domain$, $p(\env, \ctrlseq) = p(\env) \cdot p(\ctrlseq \given \env)$ where $p(\env)$ is probability of sampling $\env \in \Env$ and $p(\ctrlseq \given \env)$ is the probability of sampling $\ctrlseq \in \controlspace_{\spec(\env)}$ given $\env$. 
This distribution is key to capture the sequential nature of sampling $\ctrlseq$ only after sampling $\env$. 
For $\env \in \Env$ where $\controlspace_{\spec(\env)} = \emptyset$, $p(\ctrlseqnull \given \env) = 1$ since $\domain$ has only a single entry for $\env$, i.e, $(\env, \ctrlseqnull)$. 
In Algorithm~\ref{alg:solution}, in Line 2, we sample $\env_i \sim p(\env)$. In Line 5, we sample $\ctrlseq_i \sim p(\ctrlseq \given \env_i)$.

\begin{proposition} \label{prop:main_result_2}
Let $\domain$ be the joint sample space as defined in~\eqref{eqn:domain}, with the probability distribution $p_{\domain}= p(\env, \ctrlseq)$.
Select a `violation parameter' $\epsilon \in (0, 1)$ and a `confidence parameter' $\beta \in (0, 1)$. Pick $N$ such that
\begin{equation} \label{eqn:num_samples}
    N \geq \frac{2}{\epsilon}\left(\ln{\frac{1}{\beta}} + 1\right),
\end{equation}
then, with probability at least $1 - \beta$, the solution $\distMetricEst$ to Algorithm~\ref{alg:solution} satisfies the following conditions:
\begin{enumerate}
    \item $\prob((\env, \ctrlseq) \in \domain: \dist(\traj_\sys(\cdot; \state_0, \ctrlseq), \traj_{\abs}(\cdot; \state_0, \ctrlseq)) > \distMetricEst) \leq \epsilon$
    \item $\prob\left((\env, \ctrlseq) \in \domain: \traj_{\abs}(\cdot; \state_0, \ctrlseq) \models \spec(\env; \distMetricEst) \rightarrow \traj_{\sys}(\cdot; \state_0, \ctrlseq) \models \spec(\env)\right) > 1-~\epsilon$ provided $\controlspace_{\spec(\env, \distMetricEst)} \subseteq \controlspace_{\spec(\env)}$.
\end{enumerate}
\end{proposition} 
\noindent The proof of Proposition~\ref{prop:main_result_2} can be found in the Appendix.
Intuitively, Proposition~\ref{prop:main_result_2} states that $\distMetricEst$ is a high confidence estimate of $\distMetricNew$, if a large enough $N$ is chosen.
If we discard the confidence parameter $\beta$ for a moment, this proposition states that the size of the violation set (the set of $(\env, \ctrlseq) \in \domain$ where the corresponding distance is greater than $\distMetricEst$) is smaller than or equal to the prescribed $\epsilon$ value. 
As $\epsilon$ tends to zero, $\distMetricEst$ approaches the desired optimal solution $\distMetricNew$. 
In turn, the simulation effort grows unbounded since $N$ is inversely proportional to $\epsilon$.

As for the confidence parameter $\beta$, one should note that $\distMetricEst$ is a random quantity that depends on the randomly extracted $(\env, \ctrlseq)$ pairs. 
It may happen that the extracted samples are not representative enough, in which case the size of the violation set will be larger than $\epsilon$.
Parameter $\beta$ controls the probability that this happens; and the final result holds with probability $1-\beta$. 
Since $N$ in \eqref{eqn:num_samples} depends logarithmically on $1/\beta$; $\beta$ can be pushed down to small values such as $10^{-16}$, to make $1-\beta$ so close to $1$ to lose any practical importance.

Finally, once we have a high confidence estimate of $\distMetricNew$, we can use it with Proposition~\ref{prop:main_result_1} to provide guarantees on the safety of a controller for the system, provided that it is safe for the abstraction. (Statement (2) in Proposition~\ref{prop:main_result_2})

Note that the controller $\ctrlseq_i$ is extracted randomly from the set $\controlspace_{\spec(\env_i)}$ (Line 5). 
Obtaining $\controlspace_{\spec(\env_i)}$ and randomly sampling from it can be challenging in itself depending on the control scheme, $\Pi$, and the specification, $\spec(\env_i)$.
However, one way to randomly extract $\ctrlseq_i$ is using rejection sampling, i.e., we randomly sample controllers from the set $\controlspace_{\Pi}$ until we find a controller that satisfies the specification for the model. 
Since the controller performance is evaluated only on the model during this process, it is often cheap and does not put the system at risk. 
Nevertheless, choosing a good control scheme makes this process more efficient, as the number of samples rejected before a feasible controller is found will be fewer (see Section \ref{Sec:example_distComp} for further discussion on this).
Rejection sampling, however, poses a problem when $\controlspace_{\spec(\env_i)} = \emptyset$ and there is no way of knowing that beforehand.
In such cases, one can impose a limit on the number of rejected samples to make sure the algorithm terminates. 
This problem can also be overcome easily when there is a single safe controller for each environment, i.e., $\controlspace_{\spec(\env_i)}$ is a singleton set (see Remark 1).
%
\begin{algorithm}[t]
	\DontPrintSemicolon
	\caption{Scenario optimization for estimating \metName{}}
	\label{alg:solution}
	set $\distMetricEst = 0$ \;
	extract $N$ realizations of the environment $\env_i, i = 1, 2, \ldots, N$ \;
	\For{\text{$i=0:N-1$}}{
 	    \If{\text{$\controlspace_{\spec(\env_i)} \neq \emptyset$}}{
 	        extract a realization of a feasible controller $\ctrlseq_i \in \controlspace_{\spec(\env_i)}$ \;
 	        run the controller $\ctrlseq_i$ on $\sys$ and $\abs$, and obtain $\traj_{\sys}^i(\cdot)$ and $\traj_{\abs}^i(\cdot)$ \;
 	        compute $\dist_i = \dist(\traj_{\sys}^i(\cdot), \traj_{\abs}^i(\cdot))$
 	    }
 	    \Else{
 	        $\ctrlseq_i = \ctrlseqnull$ and $\dist_i = 0$
 	    }
    }
	set $\hat{\dist}_{\epsilon} = \max_{i \in \{1, 2, \ldots, N\}}\dist_i$
\end{algorithm}

\begin{remark}
Even though we have presented scenario optimization to estimate $\distMetricNew$, alternative derivative free optimization approaches such as Bayesian optimization, simulated annealing, evolutionary algorithms, and covariance matrix adaptation can be used as well. However, for many of these algorithms, it might be challenging to provide formal guarantees on the quality of the resultant estimate of the distance bound.
\end{remark}

 Algorithm~\ref{alg:solution} samples $N$ environment scenarios and corresponding controllers prior to running any executions on $\abs$ and $\sys$. Imagine at iteration $i$, we have $\dist_i > 0$; and if at iteration $(i+1)$, $\dist_{i+1} < \dist_i$, then the $(i+1)$th sample is not informative for approximating $\distMetricNew$. 
 A simple way to overcome this issue would be to consider only $\controlspace_{\spec(\env_i; \dist_i)}$ as the set of feasible controllers at the $(i+1)$th iteration; i.e., consider controllers where $\traj_{\abs}(\cdot) \models \spec(\env_i; \dist_i)$. This variant of Algorithm 1 would reduce the number of executions of the system; and ensure that each execution is informative for estimating $\distMetricNew$. 
 To implement this scheme, we would maintain a running max $\distMetricNew_{(i)}$ which contains the maximum of $\dist_i$ seen till now. In iteration $(i+1)$, instead of sampling from $\controlspace_{\spec(\env)}$ in Line 5, we sample from $\controlspace_{\spec(\env, \distMetricNew_{(i)})}$. Further, before the end of loop, in Line 7, we update $\distMetricNew_{(i+1)} = \max(\distMetricNew_{(i)}, \dist_{i+1})$. 
\section{Running Example: Distance Computation}
\label{Sec:example_distComp}
We now apply the proposed algorithm to compute $\distMetricNew$ for the setting described in Section \ref{sec:running_example:description}.
$\controlspace_{\spec(\env)}$ in this case is given as
\begin{align} \label{eqn:running_example:prop_condition}
\controlspace_{\spec(\env)} = & \{\ctrlseq \in \controlscheme(\env): \|\traj_{\abs}(\horizon; \state_0, \ctrlseq) - \state^*\|_2 < \gamma \}, \nonumber
\end{align}
where $\controlscheme(\env)$ is the set of LQR controllers (see Section \ref{sec:running_example:description}).
To illustrate the importance of the choice of distance metric, we compute two different distance metrics between $\sys$ and $\abs$: $\distMetricOld$ in \eqref{eqn:sim_metric:naive} and $\distMetricNew$ in \eqref{eqn:sim_metric:proposed}.
To compute $\distMetricNew$, we use Algorithm \ref{alg:solution}.
To compute $\distMetricOld$, we modify Algorithm \ref{alg:solution} to sample a random controller from $\controlscheme(\env)$ in Line 5 and compute $\dist_i$ using \eqref{eqn:sim_metric:naive} in Line 7.
%

According to the scenario approach with $\epsilon=0.01$ and $\beta=10^{-6}$, we extract $N=2964$ different reach-avoid scenarios (i.e., $N$ different final states to reach).
For each $\env_i, i \in \{1, 2, \ldots, 2964\}$, we obtain a feasible LQR controller $\ctrlseq_i \in \controlspace_{\spec(\env_i)}$ using rejection sampling. 
In particular, we randomly sample a penalty parameter $q$, solve the corresponding Riccati equation to obtain $LQR(q)$, and apply it on $\abs$. 
If the corresponding $\traj_{\abs}(\cdot)$ satisfies $\spec(\env_i)$, we use $\ctrlseq_i$ as our feasible controller sample; otherwise, we sample a new $q$ and repeat the procedure until a feasible controller is found.
This procedure tends to be really fast and requires simulating only $\abs$. 
A feasible controller was found within 3 samples of $q$ for all $\env_i$ in this case.
For $\distMetricOld$, we randomly sample a penalty parameter $q$ and use $LQR(q)$ as the controller.

The obtained distance metrics are $\distMetricOld = 0.43, \distMetricNew = 0.$
Since $\distMetricOld < \gamma$, it can be used to synthesize a safe controller for $\sys$; however, we can synthesize controller only for those reach-avoid scenarios where $\abs$ satisfies a much stringent specification: $\traj_{\abs}$ must reach within a ball of radius 0.07 around the target state.
Consequently, the set $\Env_{\spec}(\distMetricOld)$ is likely to be very small.
In contrast, $\distMetricNew = 0$; thus, Proposition \ref{prop:main_result_2} ensures that \textit{any} controller designed on $\abs$ that satisfies $\spec(\env)$ is guaranteed to satisfy it for $\sys$ as well.
In particular, the dynamics of $\sys$ and $\abs$ are same for the state $\state_1$, and state $\state_2$ is uncontrollable for $\sys$ and remain $0$ at all times.
Thus, any controller that reaches within a ball of radius $\gamma$ around a desired state $\state_1^*$ for $\abs$, if applied on $\sys$, also ensures that the system state reaches within the same ball.
Even though this relationship between $\sys$ and $\abs$ is unknown, $\distMetricNew$ is able to capture it only through simulations of $\sys$.
This example also illustrates that $\distMetricNew$ significantly reduces the conservativeness in \oldName{}, and does not unnecessarily contract the set of safe environments.

\section{Case Studies}
\label{sec:case_studies}
We now demonstrate how \metName{} can be used to obtain the safe set of environments and controllers for an autonomous quadrotor and an autonomous car.
In Section \ref{sec:quadrotor}, we demonstrate how \metName{} provides much larger safe sets compared to \oldName{}.
In Section \ref{sec:autonomous_car}, we demonstrate how \metName{} not only captures the differences between the dynamics of $\sys$ and $\abs$, but also other aspects of the system, in particular the sensor error, that might affect the satisfiability of a specification.
\subsection{Safe Altitude Control for Quadrotor}
\label{sec:quadrotor}
Our first example illustrates how the proposed distance metric behaves when the only difference between the system and the abstraction is the value of one parameter.
However, unlike the running example, the system and the abstraction dynamics are non-linear.
Moreover, we illustrate how \metName{} can be used in the cases where all safe controllers for $\abs$ may not be safe for $\sys$.

We use the reach-avoid setting described in~\cite{fisac2017general}, where the authors are interested in controlling the altitude of a quadrotor in an indoor setting while ensuring that it does not go too close to the ceiling or the floor, which are obstacles in our experiments.

A dynamic model of quadrotor vertical flight can be written as:
\begin{equation} 
\label{eqn:quad:model}
\begin{split}
    z(t+1) = & z(t) + Tv_z(t) \\
    v_z(t+1) = & v_z(t) + T(ku(t) + g),
\end{split}
\end{equation}
where $z$ is the vehicle's altitude, $v_z$ is its vertical velocity and $u$ is the commanded average thrust.
The gravitational acceleration is $g = -9.8 m/s^2$ and the discretization step $T$ is 0.01.
The control input $u(t)$ is bounded to $[0, 1]$.
We are interested in designing a controller for $\sys$ that ensures safety over a horizon of 100 timesteps. 
In particular, we have $\initStates = \{(z, v_z): 0.5 \leq z \leq 2.5 \wedge -3\leq v_z \leq 4\}$, $\avoidSets = \{\avoidSet(\cdot)\}$, and $\reachSets =\mathbb{R}^2$. 
The avoid set at any time $\time$ is given as $\avoidSet(t) = \{(z, v_z) \in \mathbb{R}^2: 0.5m \leq z(t) \leq 2.5m\}$.
We again assume that the dynamics in \eqref{eqn:quad:model} are unknown.
Consider an abstraction of $\sys$ with same dynamics as \eqref{eqn:quad:model} except that the value of parameter $k$ in the abstraction dynamics, $k_{\abs}$, is different.

The space of controllers $\controlscheme(\env)$ is given by all possible control sequences over the time horizon (i.e., $\controlscheme(\env) = \controlspace$.) 
For computing $\controlspace_{\spec(\env)}$, we use the Level Set Toolbox~\cite{mitchell2008flexible} that gives us both the set of initial states from which there exist a controller that will keep the $\traj_{\abs}(\cdot)$ outside the avoid set at all times (also called the reachable set), as well as the corresponding least restrictive controller.
In particular, we can apply any control when the abstraction trajectory is inside the reachable set and the safety-preserving control (given by the toolbox) when the trajectory is near the boundary of the reachable set.
For computation of the distance bounds, we sample a random controller sequence according to this safety-preserving control law.
If any initial state lies outside the reachable set, then it is also guaranteed that $\controlspace_{\spec(\env)}=\emptyset$ so we do not need to do any rejection sampling in this case.

When $k_{\abs} < k$, $\abs$ has strictly less control authority compared to $\sys$.
Thus, any controller that satisfies the specification for $\abs$ will also satisfy the specification for $\sys$, so $\Env_{\spec}(0)$ itself is an under approximation of $\Env_{\sys}$.
\metName{} is again able to capture this behavior.
Indeed, we computed an estimate for the distance bound using Algorithm \ref{alg:solution} and the obtained numbers are $\distMetricOld=0.30$ and $\distMetricNew=0$.
Note that not only is $\distMetricOld$ conservative, it may not be particularly useful in synthesizing a safe controller for $\sys$.
$\distMetricOld$ computed using Algorithm \ref{alg:solution} ensures that a safe controller designed on $\abs$ for $\spec(\env; \distMetricOld)$ is also safe for $\sys$ with high probability, only when this controller is \textit{randomly} selected from the set $\controlscheme$.
However, a random controller selected from $\controlscheme$ is unlikely to satisfy $\spec(\env; \distMetricOld)$ for $\abs$ itself, and thus nothing can be said about $\sys$ either.
Thus, it is hard to \textit{actually} compute an approximation of $\Env_{\sys}$.
In contrast, $\distMetricNew$ samples a controller from the set $\controlspace_{\spec(\env)}$ in Algorithm \ref{alg:solution}.
Therefore, to synthesize a controller, we \textit{randomly} select a controller from the set $\controlspace_{\spec(\env; \distMetricNew)}$, which is guaranteed to be safe on both $\abs$ and $\sys$ with high probability.
Therefore, it might be better to compare $\distMetricNew$ to $\distMetricIntOne$, which is defined similar to $\distMetricOld$, except the inner maximum in \eqref{eqn:sim_metric:naive} is computed over $\controlspace_{\spec(\env)}$ instead.
$\distMetricIntOne$ in this case turns out to be $0.5$.

Note that if we could instead compute the distance metrics exactly, $\distMetricIntOne \leq \distMetricOld$, since $\controlspace_{\spec(\env)} \subset \controlscheme$.
However, random sampling based estimate of $\distMetricIntOne$ can be greater than that of $\distMetricOld$ if the controllers corresponding to a large distance between the $\traj_{\sys}(\cdot)$ and $\traj_{\abs}(\cdot)$ are sparse in $\controlscheme$ compared to that in $\controlspace_{\spec(\env)}$.
\begin{figure}
    \centering
    \includegraphics[width=0.8\columnwidth]{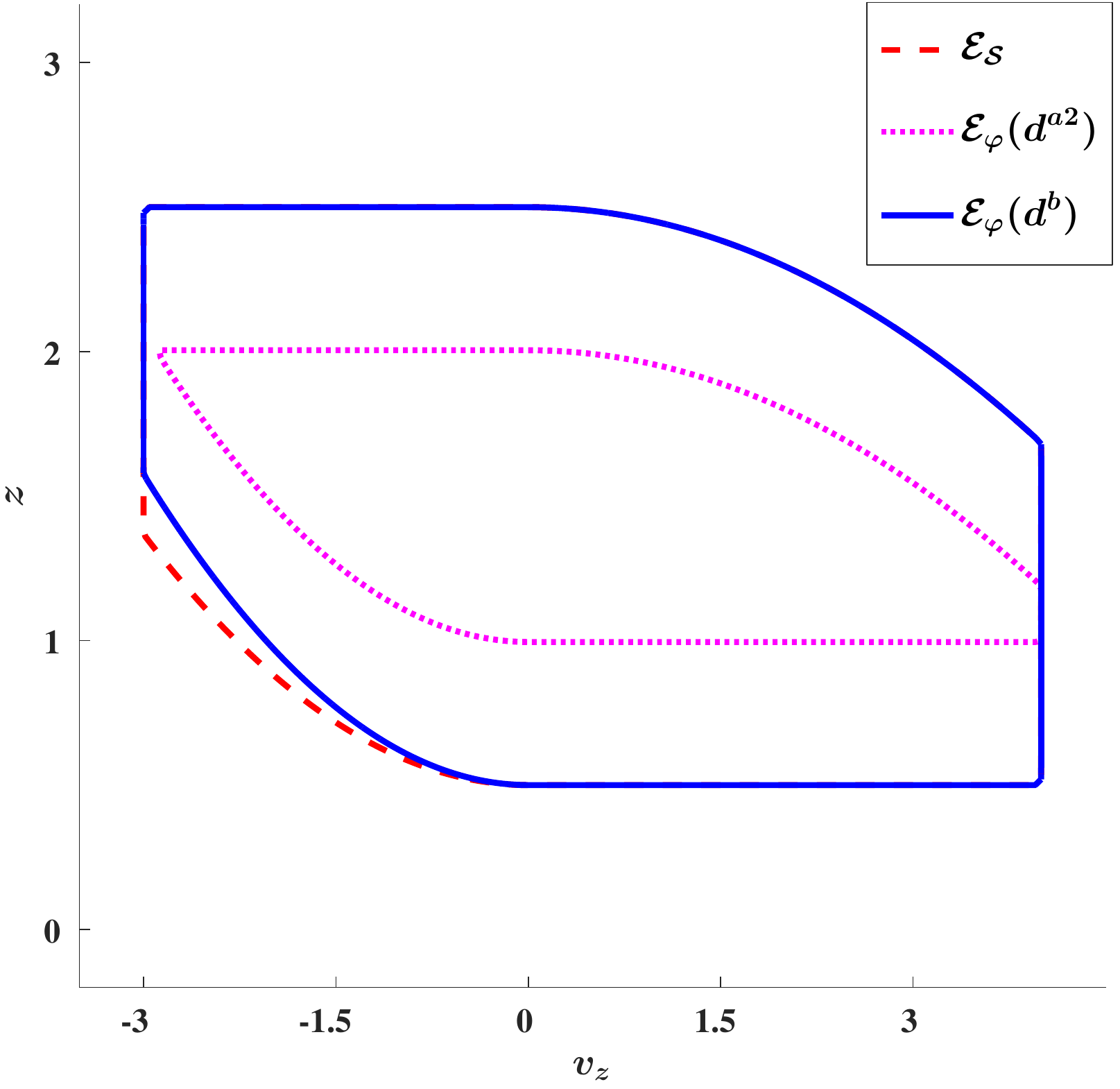}
    \caption{Different reachable sets when the quadrotor abstraction is conservative.
    The distance metric $\distMetricNew$ only considers the distance between trajectories that violates the specification on the system and satisfies it on the abstraction, leading to a less conservative estimate of the distance, and a better approximation of $\Env_{\sys}$.}
    \label{fig:quad_example:reach_sets_conservative}
\end{figure}

For illustration purposes, we also compute the reachable set $\Env_{\spec}(\distMetricNew)$, by augmenting the avoid set by $\distMetricNew$ and recomputing the reachable sets using the Level Set Toolbox.
As shown in Figure \ref{fig:quad_example:reach_sets_conservative}, $\Env_{\spec}(0)$ (the area withing the blue contour) is indeed contained within $\Env_{\sys}$ (the area within the red contour).
Here, $\Env_{\sys}$ has been computed using the system dynamics.
Even though $\Env_{\spec}(\distMetricIntOne)$ (the area within the magenta contour) is also contained in $\Env_{\sys}$, it is significantly smaller in size compared to $\Env_{\spec}(\distMetricNew)$.

\begin{figure}
    \centering
    \includegraphics[width=0.8\columnwidth]{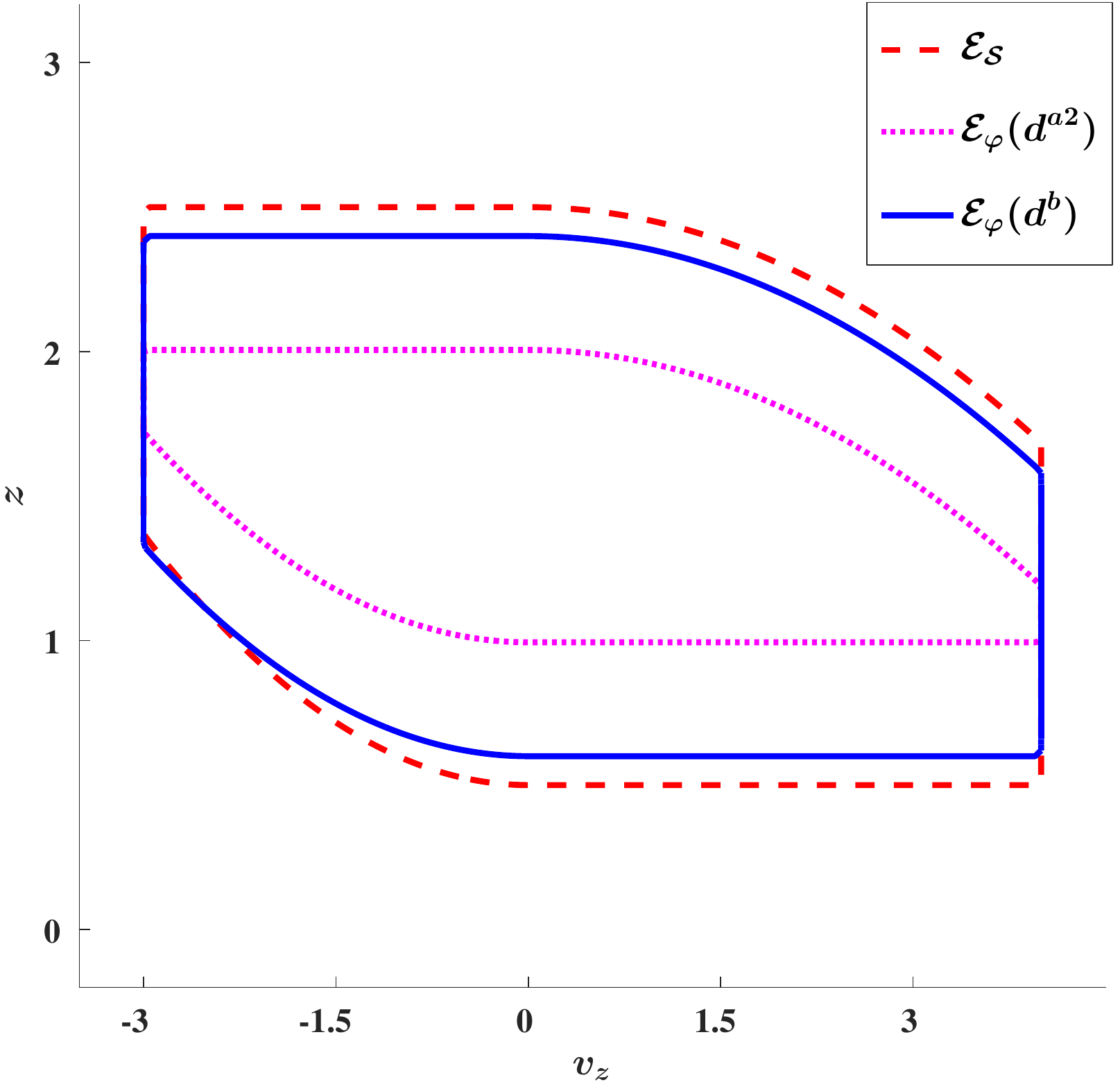}
    \caption{Different reachable sets when the quadrotor abstraction is overly optimistic.
    The distance metric $\distMetricNew$ achieves a far less conservative under-approximation of $\Env_{\sys}$ compared to the other distance metrics.}
    \label{fig:quad_example:reach_sets_optimistic}
\end{figure}

When $k_{\abs} > k$, $\sys$ has strictly less control authority compared to $\abs$.
Consequently, there might exist some environments for which it is possible to synthesize a safe controller for $\abs$, but the same controller when deployed on $\sys$ might lead to an unsafe behavior.
We again compute the distance bounds using Algorithm \ref{alg:solution} and the obtained numbers are $\distMetricOld=0.30, \distMetricIntOne=0.49, \distMetricNew=0.1$.
The corresponding reachable sets are shown in Figure \ref{fig:quad_example:reach_sets_optimistic}.
Even though we start with an overly optimistic abstraction, both $\distMetricIntOne$ and $\distMetricNew$ are able to compute an under approximation of $\Env_{\sys}$; however, the set estimated by 
$\distMetricIntOne$ is, once again, overly conservative.

\subsection{Webots: Lane Keeping}
\label{sec:autonomous_car}
We now show the application of the proposed metric for designing a safe lane keeping controller for an autonomous car. 

In this example, we use the \textbf{Webots} simulator~\cite{Webots}. The car model within the simulator is our $\sys$. For the abstraction $\abs$ we consider the bicycle model, 

\begin{minipage}{.45\linewidth}
    \centering
    \includegraphics[scale=0.5]{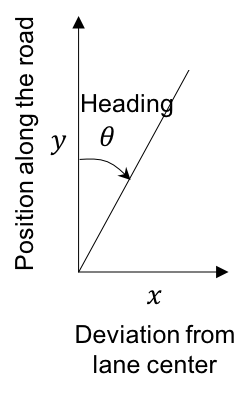}
\end{minipage}
\begin{minipage}{.45\linewidth}
\begin{equation}
    \label{eqn:bicyle_model}
    \begin{split}
        \dot{x} &= v \cdot \sin{\theta} \\
        \dot{y} &= v \cdot \cos{\theta} \\
        \dot{v} & = a \\
        \dot{\theta} & = \frac{v}{l} \tan{\omega}
    \end{split}
\end{equation}
\end{minipage}

\noindent where $[x, y, v, \theta]$ is the state, representing perpendicular deviation from the center of the lane, position along the road, speed, and heading respectively. 
The maximum speed is limited to $v_{max}=10$ km/hr.
We have two inputs, (1) a discrete acceleration control $a = \{ -\bar{a}, 0, \bar{a}\}$; and (2) a continuous steering control $\omega \in [-\pi/4, \pi/4] rad/s$. For our experiments, we use $H = 200$, which translates to about $6$ seconds of simulated trajectory.
The dynamics of $\sys$ are typically much more complex than $\abs$ and include the physical effects like friction and slip on the road.

In this case,
$\initStates = \{(x_0, \theta_0): \| x \|\leq 0.2m\, \wedge \|\theta\| \leq \pi/4 rad\}$; the initial $y_0$ and $v_0$ is set to $zero$. 
$\reachSet(t) = \{[x(t), y(t), v(t), \theta(t)] \in \mathbb{R}^4: \|x(t)\| \leq 0.5m\} \forall \time \in \timeHorizon$. 
The reach set corresponds to keeping the car within the $0.5m$ of the center of the lane.
For keeping the car in the lane, the car is equipped with two sensors, a camera (to capture the lane ahead) and compass (to measure the heading of the car). 
There is an on board perception module, which first captures the image of the road ahead; and processes it to detect the lane edges and provide an estimate of the deviation of the car from the center of the lane.

There is another car (referred to as the environment car hereon) driving in the front of $\sys$, which might obstruct the lane and cause the perception module to incorrectly detect the lane center. 
For each $\env \in \Env$, the set of possible initial states of the environment car is given by $\envParams = \{(x_e, y_e): \|x_e - x_0\|\leq 2.0m \wedge 6.25m \leq y_e - y_0 \leq 8m\}$. We set the initial speed $v_e$ and heading $\theta_e$ of the environment car to $v_{\max}$ and $0$ respectively.
We want to make sure that $\sys$ remains within the lane despite all possible initial positions of the environment car. 
For this purpose, we compute the worst-case $\distMetricNew$ across all $p \in \envParams$. 
%
%

\begin{figure}
    \centering
    \includegraphics[scale=0.5]{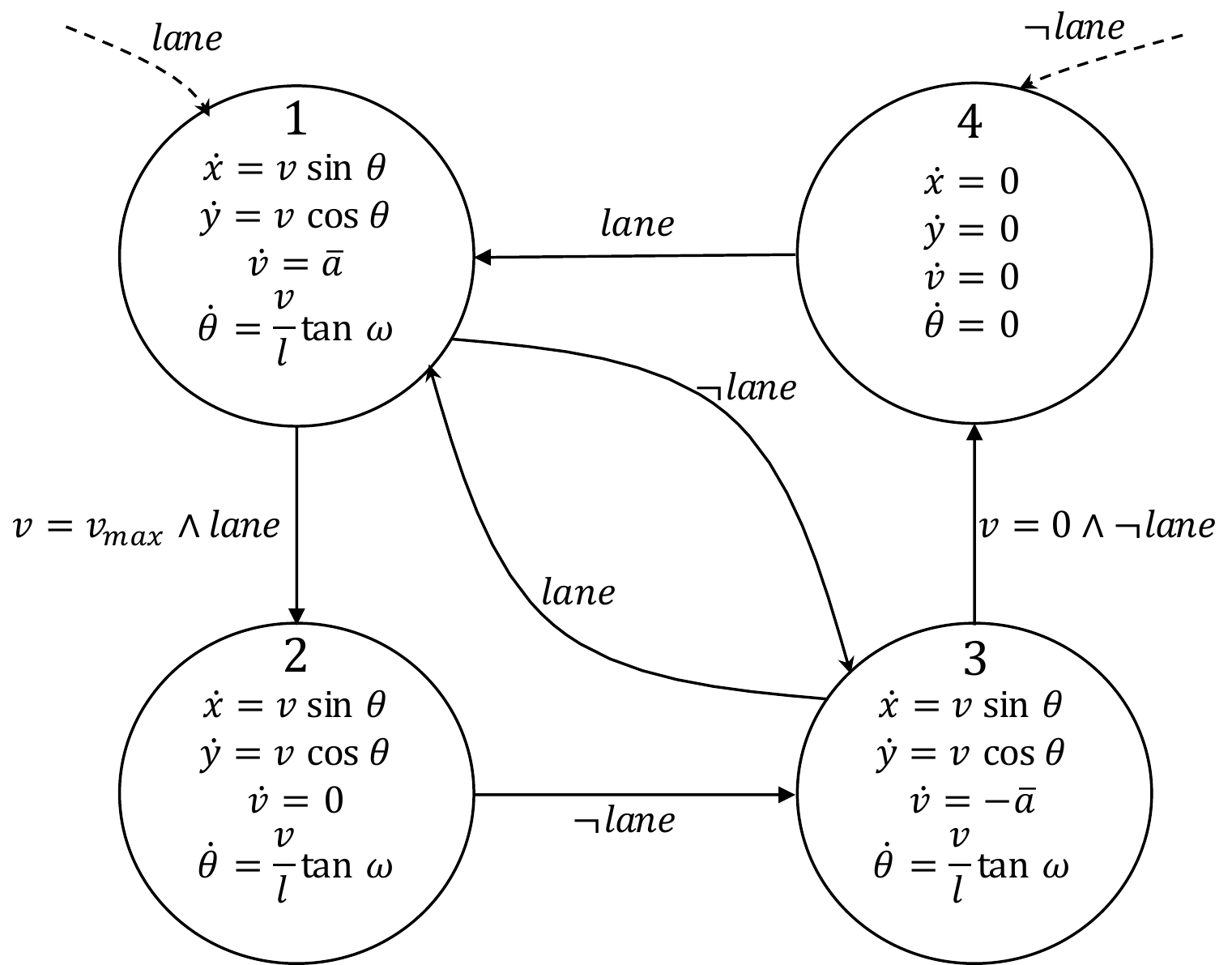}
    \caption{Hybrid controller for lane keeping. \textit{lane} means a lane is detected by the perception system. The dashed line represents the transitions taken on initialization based on the value of \textit{lane}. To closely follow the center of the lane, we synthesize a LQR controller in each mode.}
    \label{fig:hybrid_control}
\end{figure}

\begin{figure}    
    \centering
    \begin{subfigure}{0.3\textwidth}
    \includegraphics[scale=0.3]{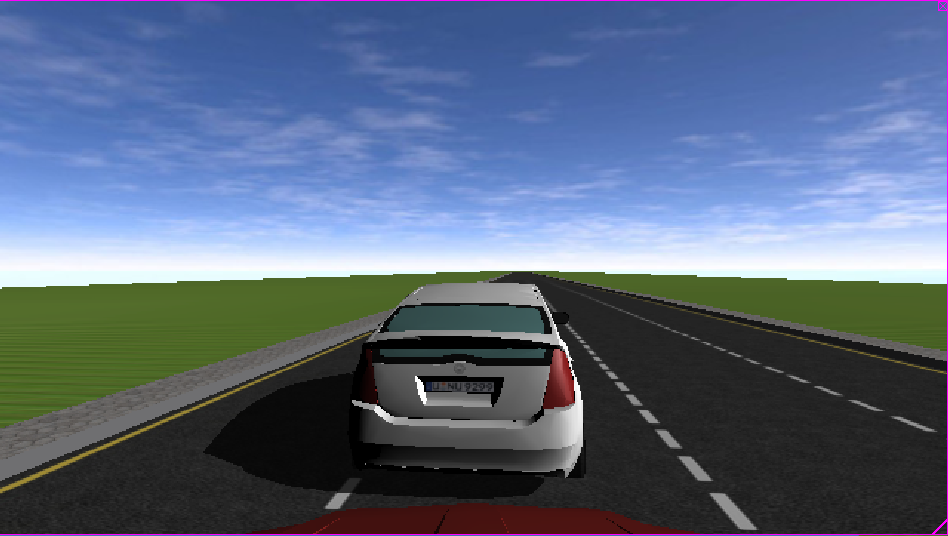}
    \caption{Environment car covers left lane.}
    \end{subfigure}
    \begin{subfigure}{0.3\textwidth}
    \includegraphics[scale=0.3]{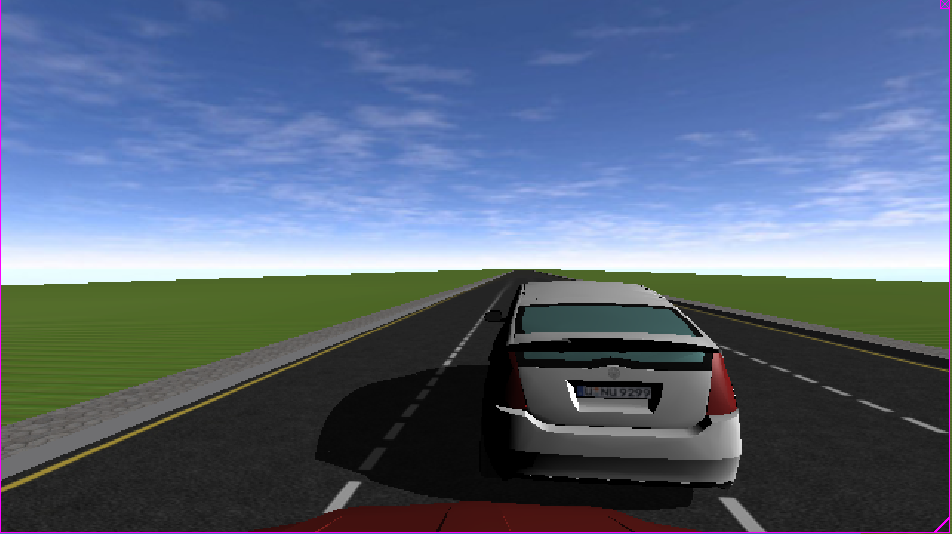}
    \caption{Shadow of environment car covers left lane.}
    \end{subfigure}
    \begin{subfigure}{0.3\textwidth}
    \includegraphics[scale=0.3]{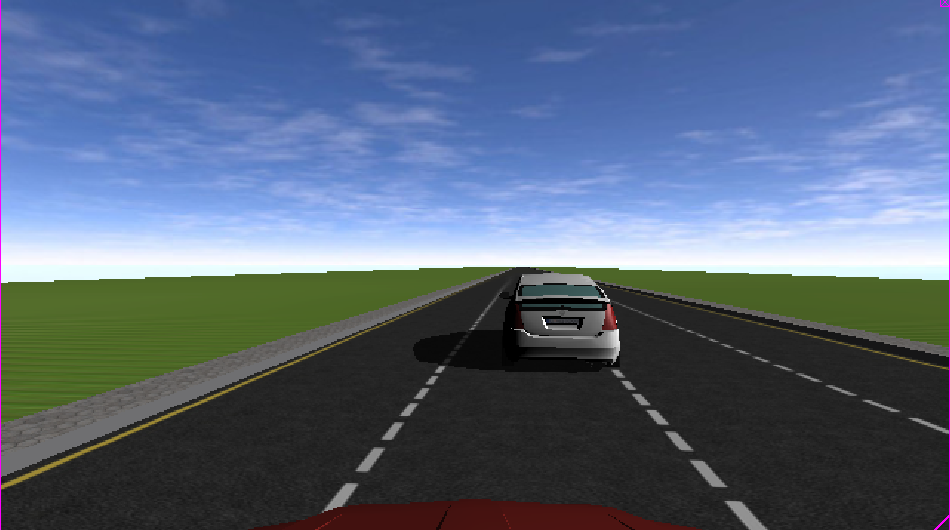}
    \caption{Lane detected correctly.}
    \end{subfigure}
    \caption{The lane detection fails for (a) and (b) and $\sys$ car tries to slow down. When lane is correctly detected (c), the LQR controller tries to follow the lane} 
    \label{fig:lane_detection}
\end{figure}

If the environment car or its shadow covers the lane edges (see Figure~\ref{fig:lane_detection} for some possible scenarios), then the lane detection fails. Technically speaking, if such a scenario occurs, then $\sys$ should slow down and come to stop until the image processing starts detecting the lane again. 
Consequently, our control scheme $\controlscheme$, is a hybrid controller shown in Figure~\ref{fig:hybrid_control}, where in each mode the controller is given by an LQR controller (with a fixed Q and R matrix) corresponding to the (linearized) dynamics in that mode. In this example, our controller is a deterministic controller since the Q and R matrices are fixed, and hence $|\controlscheme|=1$. 
In Figure~\ref{fig:hybrid_control}, in mode (1), the lane is detected and $v(t) < v_{max}$. When the $v(t) = v_{max}$ we transition to mode (2) given the lane is still detected. When the lane is no longer detected, we transition to mode (3) if $v(t) > 0$, or mode (4) if $v(t) = 0$. In modes (3) and (4), the car slows down until the lane is detected again. 
%

\begin{figure}
    \centering
    \includegraphics[scale=0.35]{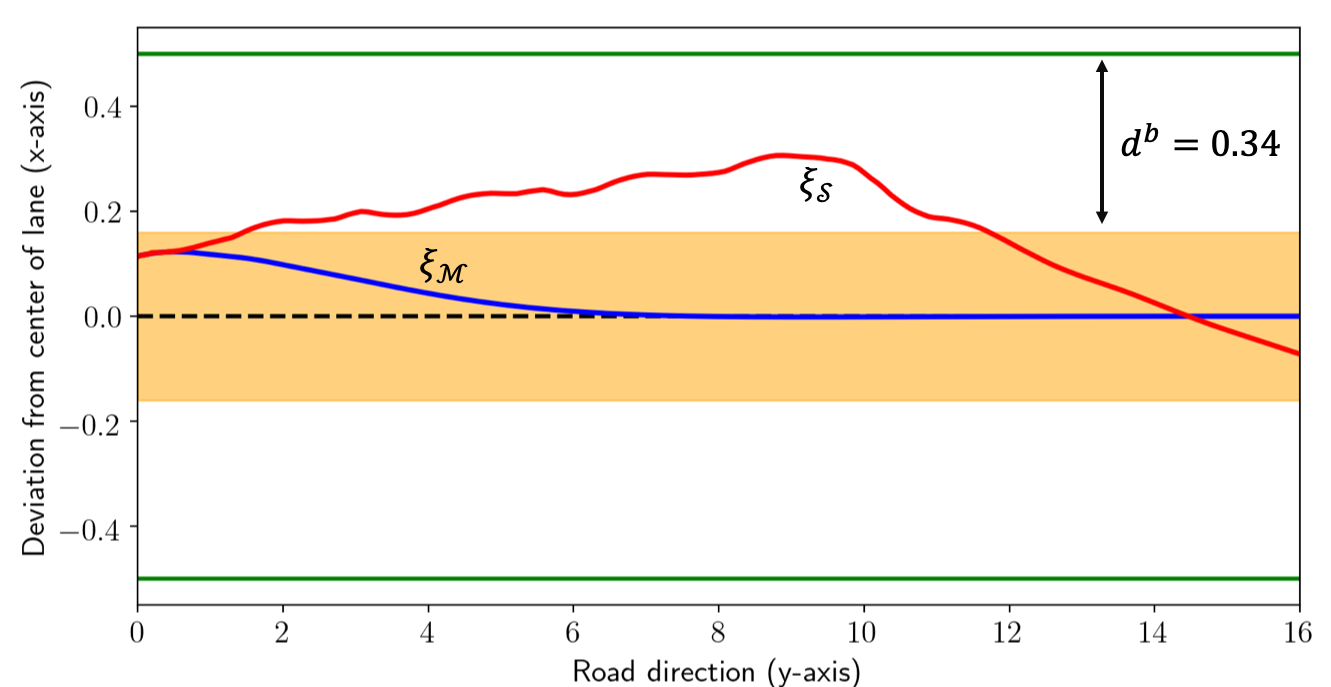}
    \caption{ The green lines represent the boundaries of the original reach set. The yellow region is the contracted reach set for the model computed using $\distMetricEst$. The model's trajectory shown in blue is entirely contained within the yellow region. Consequently, the system's trajectory (shown in dotted red) leaves the yellow region but is contained within the original reach set at all times.}
    \label{fig:safe_controller}
\end{figure}

By setting $\epsilon = 0.01$ and $\beta=1e-6$ we get $N \geq 2964$. 
We used Algorithm~\ref{alg:solution}, to sample $N$ different initial states of the $\sys$, $(x_0, \theta_0) \in \initStates$; and environment car in the simulator, $p \in \envParams$. Since the controller is deterministic, the set of feasible controllers is a singleton set, and hence we do not need to sample a feasible controller (Line 5 in Algorithm~\ref{alg:solution}).
Among these environment scenarios, the controller on $\abs$ is also able to safely control $\sys$ for 2519 scenarios.
$\distMetricEst$ is determined entirely by the remaining 445 controller, and computed to be 0.34m.
We show the application of the the computed $\distMetricEst$ for a sample environment scenario in Figure~\ref{fig:safe_controller}. The green lines represent the original reach set. The yellow shaded region represents the contracted reach set for the model computed using $\distMetricEst$. The model's trajectory (shown in blue) is contained in the yellow region and hence satisfies the more constrained specification. 
As a result, even though the system's trajectory (shown in dotted red) leaves the yellow region, it is contained within the original reach set at all times.

\begin{figure}
    \centering
    \includegraphics[scale=0.5]{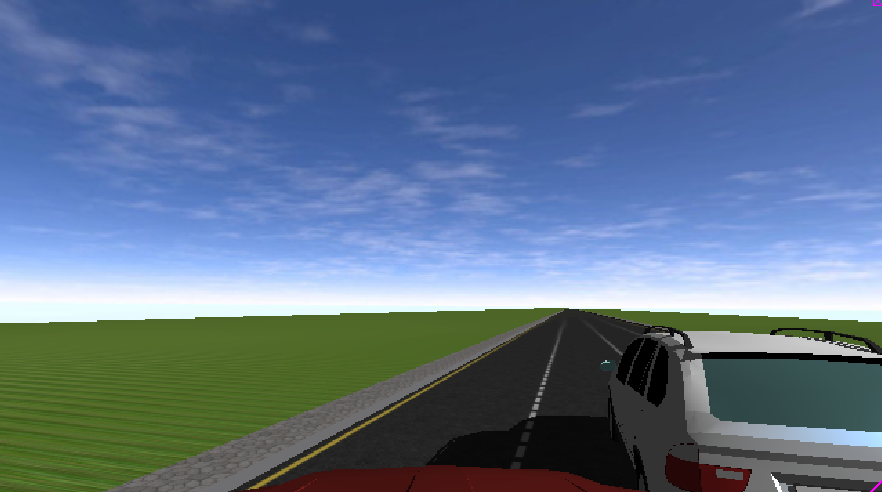}
    \caption{An example of the environment scenario that contributes to the distance between the model and the system. The environment samples used for computing \metName{} can be used to identify the reasons behind the violation of the safety specification by the system.}
    \label{fig:failed_image}
\end{figure}

We also analyze these 445 environmental scenarios that contribute to $\distMetricEst$, and notice that the fault lies within the perception module. 
In Figure~\ref{fig:failed_image}, we show one such scenario.
In this case, $\theta_0 = -\pi/4$.
Because of the left rotation of the car, the rightmost lane appears smaller and farther due to the perspective distortion.
Furthermore, the presence of the environment car completely cover the rightmost lane in the image.
The image processing module now detects the leftmost lane as the center lane and the center lane as the rightmost lane. 
Consequently, the module returns an inaccurate estimation of the center of the lane, causing $\sys$ to go outside the center lane.
This example illustrates that the samples in Algorithm \ref{alg:solution} that contributed to $\distMetricEst$ could also be used to analyze the reasons behind the violation of the safety specification by $\sys$.
%
%
%

\section{Conclusion and Future Work} \label{sec:conclusion}
Determining safe environments and synthesizing safe controllers for autonomous systems is an important problem. Typically, we rely on an abstraction of the system to synthesize controllers in different environments.
However, when a mathematical model of the system is not available, for example when the abstraction is obtained through data, it becomes challenging to provide safety guarantees for the system based on the abstraction. 
In this paper, we propose a specification-centric simulation metric \metName{} that can be used to determine the set of safe environments; and to synthesize a safe controller using such data-driven abstractions.
We also present an algorithm to compute this metric using executions on the system without knowing its true dynamics.
The proposed metric is less conservative and allows controller synthesis for reach-avoid specifications over a broader range of environments compared to the standard simulation metric.

In future, it would be interesting to extend the proposed framework for more general specifications and study its application in runtime-assurance frameworks like~\cite{herbert2017fastrack} and~\cite{desai2018}. 
Another interesting direction will be to explore active sampling methods, such as Bayesian Optimization, for the computation of \metName{}.

\bibliographystyle{ACM-Reference-Format}
\bibliography{hscc_root}

\newpage
\section{Appendix} \label{sec:appendix}
\subsection{Proof of Proposition~\ref{prop1}}
\textbf{Proof.}
Let us consider for a given environment $\env \in \Env$ and control $\ctrlseq \in  \controlscheme(\env)$, $\traj_{\abs}(\cdot; \state_0, \ctrlseq) \models \spec(\env; \distMetricOld)$.
We would like to prove that $\traj_{\sys}(\cdot; \state_0, \ctrlseq) \models \spec(\env)$.

From~\eqref{eqn:sim_metric:naive}, we have 
\begin{equation}
\label{eqn:prop1:traj_sys}
    \|\traj_{\sys}(t) - \traj_{\abs}(t)\| \leq \distMetricOld\; \forall t \in \timeHorizon. 
\end{equation}
\noindent From the definition of specification in~\eqref{eqn:spec:modified}, we have $\traj_{\abs}(\cdot) \models \spec(\env; \distMetricOld)$ if and only if $\traj_{\abs}(\cdot) \in \spec(\env; \distMetricOld)$.
Therefore, $\traj_{\abs}(t) \notin \avoidSet(t) \oplus \distMetricOld$ and $\traj_{\abs}(t) \in \reachSet(t) \ominus \distMetricOld\; \forall t \in \timeHorizon$.
%
Since $\traj_{\abs}(t) \notin \avoidSet(t) \oplus \distMetricOld$,
\begin{equation} \label{eqn:prop1:proof:help1}
    \|\traj_{\abs}(t) - a\| > \distMetricOld\;, \forall t \in \timeHorizon\;, \forall a \in \avoidSet(t).
\end{equation}
\noindent Combining \eqref{eqn:prop1:traj_sys} and \eqref{eqn:prop1:proof:help1} implies that 
\begin{equation} \label{eqn:prop1:proof:help2}
    \|\traj_{\sys}(t) - a\| > 0\;, \forall t \in \timeHorizon\;, \forall a \in \avoidSet(t).
\end{equation}
Equation \eqref{eqn:prop1:proof:help2} implies that $\traj_{\sys}(t) \notin \avoidSet(t)$ for any $ t \in \timeHorizon$. Similarly, it can be shown that 
\begin{equation*} 
    \|\traj_{\sys}(t) - r\| > 0\;, \forall t \in \timeHorizon\;, \forall r \in \reachSet(t)^c,
\end{equation*}
where $\reachSet(t)^c$ denotes the complement of the set $\reachSet(t)$. Therefore, $\traj_{\sys}(t) \in \reachSet(t)\; \forall t \in \timeHorizon$.

Since $\traj_{\sys}(t) \notin \avoidSet(t)$ and $\traj_{\sys}(t) \in \reachSet(t)$ for all $ t \in \timeHorizon$, we have $\traj_{\sys}(\cdot; \state_0, \ctrlseq) \models \spec(\env)$. 
%
\hfill $\square$

\subsection{Proof of Proposition~\ref{prop:main_result_1}}
\textbf{Proof.}
We prove the desired result by contradiction. Suppose there exists an environment $\env \in \Env$ and a controller $\ctrlseq \in \controlspace_{\spec(\env, \distMetricNew)}$ such that
$\traj_{\abs}(\cdot; \state_0, \ctrlseq) \models \spec(\env; \distMetricNew)$ but $\traj_{\sys}(\cdot; \state_0, \ctrlseq) \not\models \spec(\env)$. 
%
%

Since $\traj_{\abs}(\cdot; \state_0, \ctrlseq) \models \spec(\env; \distMetricNew)$, we have that,
\begin{align} 
    \forall t \in \timeHorizon \; \traj_{\abs}(t; \state_0, \ctrlseq) & \notin \avoidSet(t; \distMetricNew) = \avoidSet(t) \oplus \distMetricNew \label{eqn:prop2:proof:help1}\\
        \forall t \in \timeHorizon \; \traj_{\abs}(t; \state_0, \ctrlseq) &\in \reachSet(t; \distMetricNew) = \reachSet(t) \ominus \distMetricNew \label{eqn:prop2:proof:help2}
\end{align}
%
Since $\distMetricNew$ is the solution to~\eqref{eqn:sim_metric:proposed}, and $\ctrlseq \in \controlspace_{\spec(\env)}$ (as $\controlspace_{\spec(\env; \distMetricNew)} \subseteq \controlspace_{\spec(\env)}$) is such that $\traj_{\abs}(\cdot) \not\models \spec(\env)$,~\eqref{eqn:sim_metric:proposed} and ~\eqref{eqn:dist_fn:proposed} imply that,
\begin{equation}
    \min_{\time \in \timeHorizon}
    \left(\min\{\signdist\left(\traj_{\abs}(\time), \avoidSet(\time)\right),
    -\signdist\left(\traj_{\abs}(\time), \reachSet(\time)\right)
    \}\right) \leq \distMetricNew.
\end{equation}
Therefore, $\exists t'\in \timeHorizon$ such that 
\begin{equation}
    \min\{\signdist\left(\traj_{\abs}(t'), \avoidSet(t')\right),
    -\signdist\left(\traj_{\abs}(t'), \reachSet(t')\right)
    \} \leq \distMetricNew,
\end{equation}
which implies that either 
\begin{enumerate}
    \item $\signdist\left(\traj_{\abs}(t'), \avoidSet(t')\right) \leq \distMetricNew$, or
    \item $\signdist\left(\traj_{\abs}(t'), \reachSet(t')\right) \geq -\distMetricNew$
\end{enumerate}
%

If $\signdist\left(\traj_{\abs}(t'), \avoidSet(t')\right) \leq \distMetricNew$, $\exists a \in \avoidSet(t')$ such that
\begin{equation*}
    \|\traj_{\abs}(t') - a\| \leq \distMetricNew.
\end{equation*}
Therefore, $\traj_{\abs}(t') \in \avoidSet(t') \oplus \distMetricNew$, which contradicts \eqref{eqn:prop2:proof:help1}.
Similarly, if $\signdist\left(\traj_{\abs}(t'), \reachSet(t')\right) \geq -\distMetricNew$, $\exists r \in \reachSet(t')^c$ such that
\begin{equation*}
    \|\traj_{\abs}(t') - r\| \leq \distMetricNew,
\end{equation*}
which implies that $\traj_{\abs}(t') \not\in \reachSet(t') \ominus \distMetricNew$, which contradicts \eqref{eqn:prop2:proof:help2}.

When $\controlspace_{\spec(\env, \distMetricNew)} \not\subseteq \controlspace_{\spec(\env)}$, for any controller $\ctrlseq \in \controlspace_{\spec(\env, \distMetricNew)} \setminus \controlspace_{\spec(\env)}$ such that $\traj_{\abs}(\cdot; \state_0, \ctrlseq) \models \spec(\env; \distMetricNew)$, we can no longer comment on the behavior of the corresponding system trajectory. 
This is because while computing $\distMetricNew$, these controllers were not taken into account. 
%
%
%
\hfill $\square$

\subsection{Proof of Proposition~\ref{prop:main_result_2}}
\textbf{Proof. }
Statement (1) of Proposition~\ref{prop:main_result_2} follows directly from the guarantees provided by Scenario Optimization (Theorem 1 in~\cite{campi2009}). To use the result in \cite{campi2009}, we need to prove: (a) computing $\distMetricNew$ can be converted into a standard Scenario Optimization problem and (b) Algorithm~\ref{alg:solution} samples i.i.d from $\domain$ with probability $p_{\domain}$.

\eqref{eqn:sim_metric:proposed} can be re-written as $\distMetricNew = \max_{(\env, \ctrlseq) \in \domain} \dist(\traj_{\sys}(\cdot), \traj_{\abs}(\cdot))$ which can be formalized as the following optimization problem,
\begin{equation*}
    \begin{split}
        \min \; &g \\
        \text{s.t.} \;& \forall (\env, \ctrlseq) \in \domain \;,  \dist(\traj_{\sys}(\cdot), \traj_{\abs}(\cdot)) \leq g
    \end{split}
\end{equation*}
This is semi-infinite optimization problem where the constraints are convex (in fact, linear) in the optimization variable $g$ for any given $(\env, \ctrlseq)$. 
Statement (1) now follows from Theorem 1 in~\cite{campi2009} by replacing $c=1$, $\gamma$ by $g$, $\Delta$ by $\domain$, and $f$ by $\dist(\traj_{\sys}(\cdot), \traj_{\abs}(\cdot)) - g$.
Theorem 1 in~\cite{campi2009}, however, requires that i.i.d samples are chosen from the distribution $p_{\domain}$.
This can be proved by noticing that, in Algorithm~\ref{alg:solution}, we first sample $\env_i \sim p(\env)$ (in Line 2), and then sample $\ctrlseq_i \sim p(\ctrlseq \given \env)$ (in Line 4.) 
Hence, every $(\env_i, \ctrlseq_i)$ is sampled from $p_{\domain} = p(\env) \cdot p(\ctrlseq \given \env)$. Since each $i = 1, \dots, N$ is sampled randomly and independent of each other, the $(\env_i, \ctrlseq_i)$ pairs are indeed sampled i.i.d from $p_{\domain}$.

Algorithm~\ref{alg:solution} returns an estimate $\distMetricEst$ for $\distMetricNew$. We have already established that $\distMetricEst$ satisfies the probabilistic guarantees provided by scenario optimization (Statement (1)). 
From Proposition~\ref{prop:main_result_1}, we have $\forall (\env, \ctrlseq) \in \domain$ where $\dist(\traj_{\sys}(\cdot), \traj_{\abs}(\cdot)) \leq \distMetricEst$, $\traj_{\abs}(\cdot; \state_0, \ctrlseq) \models \spec(\env; \distMetricEst) \rightarrow \traj_{\sys}(\cdot; \state_0, \ctrlseq) \models \spec(\env)$, provided $\controlspace_{\spec(\env;\distMetricEst)} \subseteq \controlspace_{\spec(\env)}$. 
Therefore,
\begin{equation*}
    \prob\left((\env, \ctrlseq) \in \domain: \traj_{\abs}(\cdot; \state_0, \ctrlseq) \models \spec(\env; \distMetricEst) \rightarrow \traj_{\sys}(\cdot; \state_0, \ctrlseq) \models \spec(\env)\right) > 1-\epsilon. 
\end{equation*}
%
%
\hfill $\square$

\subsection{Proof of Theorem~\ref{thm:min_dist} and Corollary~\ref{cor:spec_containment}}
\textbf{Proof. } Consider any $d > \distMetricNew$. From the statement of Theorem~\ref{thm:min_dist}, we have that $\controlspace_{\spec(\env; d)} \subseteq \controlspace_{\spec(\env; \distMetricNew)}$. Hence, $\Env_{\spec}(d) \subseteq \Env_{\spec}(\distMetricNew)$ follows from the definition of $\Env_{\spec}(\dist)$ in~\eqref{eqn:controller:modified}.
$\controlspace_{\spec(\env; \distMetricNew)} \subseteq \controlspace_{\sys}(\env)$ and $\Env_{\spec}(\distMetricNew) \subseteq \Env_{\sys}$ is already ensured by Proposition~\ref{prop:main_result_1}, and hence Theorem 1 follows.

We now prove that for all $0 < d < \distMetricNew$, $\exists\; \env \in \Env$ such that~\eqref{eqn:theorem1:condition} does not hold, and hence the result of Theorem 1 trivially holds.
We prove the result by contradiction. 
Suppose $0 < d < \distMetricNew$ be such that~\eqref{eqn:theorem1:condition} holds.
Let $(\env^*, \ctrlseq^*)$ be the environment, controller pair where $\dist(\traj_{\abs}(\cdot; \state_0^*,\ctrlseq^*), \traj_{\sys}(\cdot; \state_0^*, \ctrlseq^*))~=~\distMetricNew$.
Equation~\eqref{eqn:sim_metric:proposed} and ~\eqref{eqn:dist_fn:proposed} thus imply that
\begin{equation} \label{eqn:theorem1:proof:help1}
    \min_{\time \in \timeHorizon}
    \left(\min\{\signdist\left(\traj_{\abs}(\time; \state_0^*, \ctrlseq^*), \avoidSet^*(\time)\right),
    -\signdist\left(\traj_{\abs}(\time; \state_0^*, \ctrlseq^*), \reachSet^*(\time)\right)
    \}\right) = \distMetricNew,
\end{equation}
and $\traj_{\sys}(\cdot; \state_0^*, \ctrlseq^*) \not\models \spec(\env^*)$. Equation \eqref{eqn:theorem1:proof:help1} implies that 
\begin{align}
     \forall \time \in \timeHorizon,\; & \signdist\left(\traj_{\abs}(\time; \state_0^*, \ctrlseq^*), \avoidSet^*(\time)\right) \geq \distMetricNew  \label{eqn:theorem1:proof:help2} \\
          \forall \time \in \timeHorizon,\; & \signdist\left(\traj_{\abs}(\time; \state_0^*, \ctrlseq^*), \reachSet^*(\time)\right) \leq -\distMetricNew.  \label{eqn:theorem1:proof:help3}
\end{align}
Equations \eqref{eqn:theorem1:proof:help2} and \eqref{eqn:theorem1:proof:help3} imply that
\begin{equation} \label{eqn:theorem1:proof:help4}
\forall \time \in \timeHorizon,\; \traj_{\abs}(\time; \state_0^*, \ctrlseq^*) \not\in \avoidSet^*(t) \oplus d, \traj_{\abs}(\time; \state_0^*, \ctrlseq^*) \in \reachSet^*(t) \ominus d.
\end{equation}
Consequently, we have $\traj_{\abs}(\cdot; \state_0^*, \ctrlseq^*) \models \spec(\env^*; d)$.
This contradicts~\eqref{eqn:theorem1:condition} since $\traj_{\sys}(\cdot; \state_0^*, \ctrlseq^*) \not\models \spec(\env^*)$.
Therefore, for all $0 < d < \distMetricNew$, $\exists\; \env \in \Env\;, \ctrlseq \in \controlscheme(\env) \;, \traj_{\abs}(\cdot)\models \spec(\env;d) \not\rightarrow \traj_{\sys}(\cdot)\models \spec(\env)$.

To prove the corollary, we first prove that if $d_1 > d_2$, then $\traj_{\abs}(\cdot) \models \spec(\env; d_1)$ implies $\traj_{\abs}(\cdot) \models \spec(\env; d_2)\;, \forall \env \in \Env$.
Since $\traj_{\abs}(\cdot) \models \spec(\env; d_1)$, we have
\begin{equation*}
    \forall \time \in \timeHorizon, \traj_{\abs}(t) \notin \avoidSet(t) \oplus d_1,  \traj_{\abs}(t) \in \reachSet(t) \ominus d_1
\end{equation*}
Since $d_1 > d_2$, the above equation implies that
\begin{equation*}
    \forall \time \in \timeHorizon, \traj_{\abs}(t) \notin \avoidSet(t) \oplus d_2,  \traj_{\abs}(t) \in \reachSet(t) \ominus d_2.
\end{equation*}
Therefore, $\traj_{\abs}(\cdot) \models \spec(\env; d_2)$.
The corollary now follows from noting that for all $0 < d < \distMetricNew$, $\exists\; \env \in \Env\;, \ctrlseq \in \controlscheme(\env) \;, \traj_{\abs}(\cdot)\models \spec(\env;d) \not\rightarrow \traj_{\sys}(\cdot)\models \spec(\env)$.
\hfill $\square$

\end{document}